\documentclass[manuscript,screen,nonacm]{acmart}
\usepackage{gensymb,graphics,subfigure,inputenc,setspace}
\usepackage[algo2e,boxed]{algorithm2e}
\usepackage[figuresright]{rotating}
\usepackage{textcomp,subfigure,multirow,upgreek}
\usepackage{booktabs}
\usepackage{mdwlist}

\usepackage{ragged2e}
\usepackage{bm}
\usepackage{fontawesome}
\usepackage{wrapfig}
\graphicspath{{Figures/}}

\usepackage{caption}
\usepackage{enumitem}
\setitemize{noitemsep,topsep=0pt,parsep=0pt,partopsep=0pt}
\AtBeginDocument{%
  \providecommand\BibTeX{{%
    \normalfont B\kern-0.5em{\scshape i\kern-0.25em b}\kern-0.8em\TeX}}}

\setcopyright{acmcopyright}
\acmJournal{TOMM}
\acmYear{2022} \acmVolume{1} \acmNumber{1} \acmArticle{1} \acmMonth{1} \acmPrice{15.00}\acmDOI{10.1145/3513134}

\begin{document}

\title{Disentangle Saliency Detection into Cascaded Detail Modeling and Body Filling}

\author{Yue Song}
\authornote{Both authors contributed equally to this research.}
\email{yue.song@unitn.it}

\affiliation{%
  \institution{University of Trento}
  \country{Italy}
  \postcode{38122}
}

\author{Hao Tang}
\authornotemark[1]
\email{hao.tang@vision.ee.ethz.ch}

\affiliation{%
  \institution{ETH Zurich}
  \country{Switzerland}
  \postcode{8092}
}

\author{Nicu Sebe}
\email{nicu.sebe@unitn.it}

\affiliation{%
  \institution{University of Trento}
  \country{Italy}
  \postcode{38122}
}

\author{Wei Wang}
\email{wei.wang@unitn.it}

\affiliation{%
  \institution{University of Trento}
  \country{Italy}
  \postcode{38122}
}

\renewcommand{\shortauthors}{Song and Tang, et al.}

\begin{abstract}
Salient object detection has been long studied to identify the most visually attractive objects in images/videos. Recently, a growing amount of approaches have been proposed all of which rely on the contour/edge information to improve detection performance. 
The edge labels are either put into the loss directly or used as extra supervision. The edge and body can also be learned separately and then fused afterward.
Both methods either lead to high prediction errors near the edge or cannot be trained in an end-to-end manner. Another problem is that existing methods may fail to detect objects of various sizes due to the lack of efficient and effective feature fusion mechanisms.
In this work, we propose to decompose the saliency detection task into two cascaded sub-tasks, \emph{i.e.}, detail modeling and body filling.
Specifically, the detail modeling focuses on capturing the object edges by supervision of explicitly decomposed detail label that consists of the pixels that are nested on the edge and near the edge.
Then the body filling learns the body part which will be filled into the detail map to generate more accurate saliency map. 
To effectively fuse the features and handle objects at different scales, we have also proposed two novel multi-scale detail attention and body attention blocks for precise detail and body modeling. 
Experimental results show that our method achieves state-of-the-art performances on six public datasets.
\end{abstract}

\ccsdesc[500]{Computing methodologies~Scene understanding}

\keywords{Salient Object Detection, Visual Saliency, Foreground Segmentation}

\maketitle

\section{Introduction}
Human Visual System (HVS) has the innate ability to capture salient objects from visual scenes rapidly without training~\cite{treisman1980feature}. Salient Object Detection (SOD) aims at simulating HVS to detect distinctive regions or objects, where people would like to focus their eyes on~\cite{borji2015salient,wang2021salient}. In the past decades, it has attracted much interest from research communities, mainly because it can find objects or regions that can represent a scene efficiently, a useful step in downstream computer vision tasks. Saliency detection models have been evolved from traditional hand-engineering approaches via different saliency cues (\emph{e.g.}, global contrast~\cite{cheng2014global}, background prior~\cite{yang2013saliency}, and spectral analysis~\cite{hou2007saliency}) to Fully Convolutional Neural Networks (FCN)~\cite{long2015fully} based methods.

\begin{wrapfigure}{r}{0.5\linewidth}
    \centering
    \includegraphics[width=0.99\linewidth]{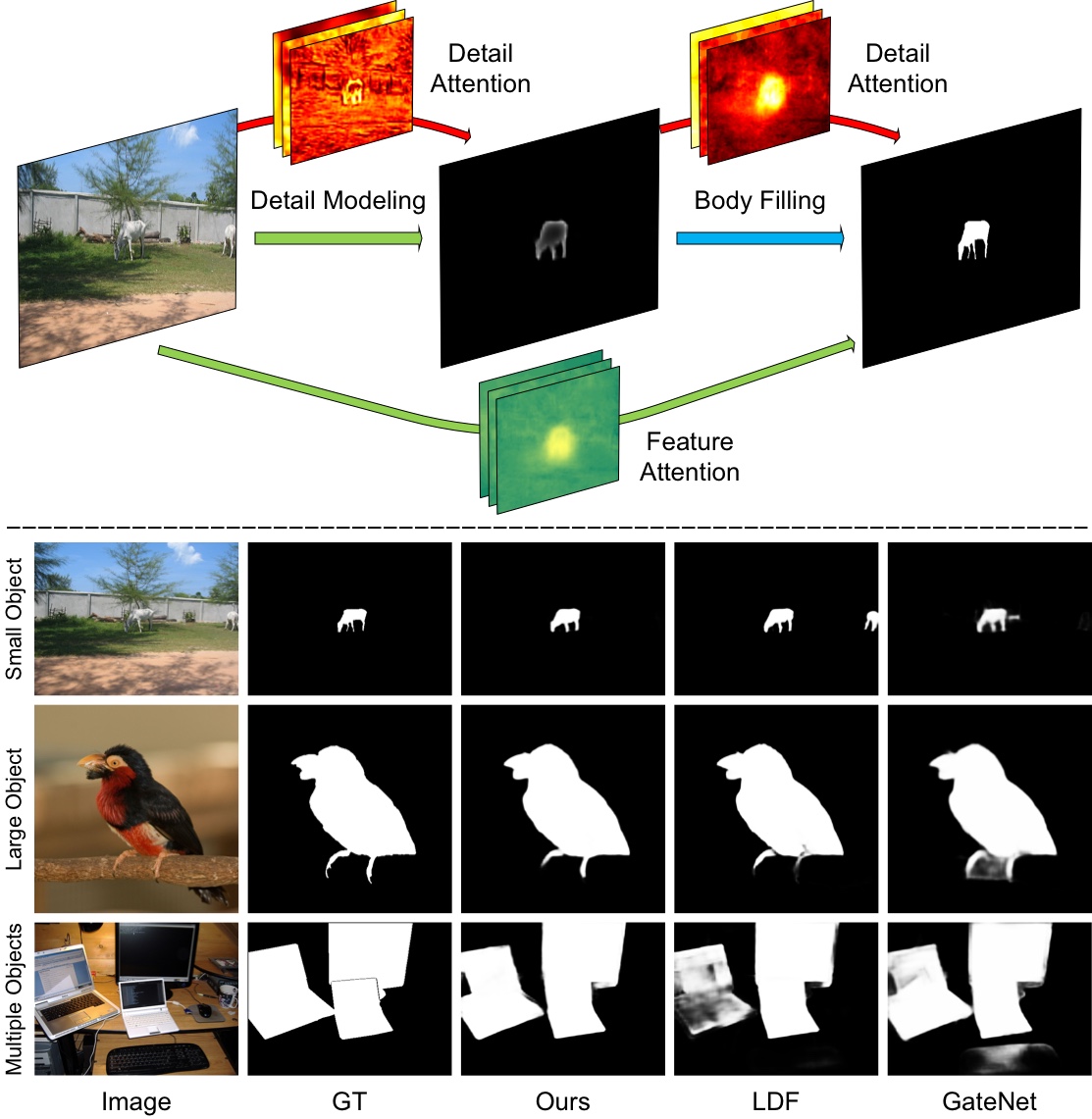}
    \caption{(\textbf{Top}) We disentangle the task of salient object detection into cascaded detail modeling and body filling. The proposed multi-scale attention blocks polish the features passed via short connections and help the network attend to the salient regions. (\textbf{Bottom}) Qualitative comparison between our method and two recent state-of-the-art methods LDF~\cite{wei2020label} and GateNet~\cite{zhao2020suppress}. Our approach can precisely segment objects of various sizes with subtle details.}
    \label{fig:story}
\end{wrapfigure}

Despite that FCN-based solutions~\cite{hou2017deeply,liu2018picanet,zhang2018bi,zhao2019egnet,qin2019basnet,su2019selectivity,wei2020f3net,zhao2020suppress,pang2020multi,wei2020label} have made remarkable progress so far, there still exist two main challenges: (i) the pixels near the object edge have a very imbalanced distribution, which makes these pixels harder to predict than the non-edge ones. Existing saliency detection models usually get large prediction error when the pixel is close to the object boundary~\cite{wei2020label}; (ii) most saliency detection methods build models on the encoder-decoder framework and develop different strategies to aggregate multi-scale features for better representation. However, due to the lack of effective fusion mechanisms to integrate multi-scale or multi-level feature, the generated saliency maps may fail to accurately predict objects in different scales. Because of these two issues, existing methods might fail to generate accurate saliency maps with sharp boundaries and coherent details (see Fig.~\ref{fig:story}).

For the first problem, many methods attempted to introduce boundary information as extra supervision to improve the prediction performances~\cite{zhao2019egnet,qin2019basnet,su2019selectivity,liu2019simple,feng2019attentive,wu2019stacked}. However, the introduced edge label only indicates the pixel on the edge. Its direct use as supervision can decrease the global saliency prediction error but will degrade the prediction performance near the edge~\cite{wei2020label}. More recently, Wei~\emph{et al.}~\cite{wei2020label} proposed to explicitly decompose the ground truth saliency label into the body label and the detail label. The decomposed detail label consists of both edges as well as nearby pixels, which makes full use of pixels near the edge and thus has a more balanced pixel distribution. The decoupled body and detail maps are used to train two separate network branches, and a feature fusion branch is needed to combine the two streams to generate the final saliency map. Their proposed architecture involves two iterations \emph{i.e.}, train the detail and body branches until the two branches can output good body/detail maps, and then train the fusion module. It is non-trivial to control the two iterations and train the model in an end-to-end manner.

For the second problem, some methods tried to pass the features at the corresponding level in the encoder to the decoder via different connection pathways to leverage multi-level context information~\cite{luo2017non,zhang2018bi,zhang2018progressive,chen2018reverse,feng2019attentive,wu2019mutual,wu2019cascaded,wang2021salient,qin2019basnet,zhong2021highly}. Without being processed by proper mechanism, the representation power of details in a single shallow layer may be weakened or disturbed by deeper features with high-level semantic information. To more effectively utilize multi-scale features, some methods proposed to pass multi-layer features to a decoder in a single layer in the fully connected manner or the heuristic style~\cite{zhang2017amulet,hou2017deeply,wang2018detect}. However, this kind of solution suffers from huge computational burden and fusion difficulties brought by the excessive amount of features and their resolution gap. There lacks such a mechanism that can effectively and efficiently fuse multi-level features without losing representation power at different scales.

To address both aforementioned issues, we propose a novel solution: a cascaded framework that disentangles traditional SOD task into two sub-tasks, where the first sub-network is forced to generate the detail map by supervision of decomposed detail label as proposed in~\cite{wei2020label}, subsequently the second sub-network takes in the detail map, detail feature, and the image to generate the body map and fuse into the final saliency map. The proposed framework explicitly divides the original task into two cascaded sub-tasks each of which has its own specific target. This reduces the difficulty in directly predicting the whole saliency map. Besides the framework, we also propose two novel multi-scale attention blocks that aim at fusing features at different levels and detecting objects of various sizes. The proposed blocks can enrich the fused feature with multi-scale attention, which can effectively fuse two or three multi-level features and relax the difficulty in the detection of multi-scale objects. In addition, we suggest a hybrid loss setting that targets the accurate generation of each map and can well complement each other. Our model is trained in an end-to-end fashion and has a reasonable inference speed of 20 FPS on a single GPU. The proposed model is thoroughly validated under four metrics across six public benchmark datasets to demonstrate its superior performances. 

In summary, the main contributions of the paper are as follows:
\begin{itemize}
    \item We propose a novel cascaded saliency detection framework that first produces detail maps of the object and then generates accurate saliency map by filling the detail map with body map. The proposed framework reduces the difficulty in directly predicting the whole saliency map and can be trained efficiently in an end-to-end manner.
    \item We propose two novel multi-scale attention blocks that can attentively fuse multiple features at multiple scales for precise detail and body map generation.
    We also suggest a hybrid loss setting that specifically targets the detail and body maps and complement each other.
    \item Our proposed model achieves state-of-the-art performances against 10 most recent state-of-the-art methods on six benchmark datasets under four widely used metrics. 
    Extensive ablation studies are also conducted to demonstrate the effectiveness of each proposed module.
\end{itemize}
\section{Related Work}

Early saliency detection methods in hand-engineered era mainly rely on various saliency cues, including global or local contrast~\cite{cheng2014global,felzenszwalb2004efficient}, background prior~\cite{yang2013saliency}, and spectral analysis~\cite{hou2007saliency,guo2008spatio}. Due to the page limits, the readers are kindly referred to~\cite{borji2015salient} for a detailed review. Here we recap modern approaches in deep learning era. These FCN-based methods can be broadly divided into two families as follows:

\noindent \textbf{Aggregation-based Models.}
Most modern saliency detection models are based on the encoder-decoder framework to integrate multi-level features and leverage contextual information across different layers~\cite{tong2015salient,zhang2017salient,luo2017non,zhang2018bi,zhang2018progressive,chen2018reverse,feng2019attentive,wu2019mutual,wu2019cascaded,qin2019basnet,zhang2017amulet,hou2017deeply,wang2018detect,wei2020f3net,pang2020multi,bruno2020multi}. The encoder is often used to extract multi-level features from the image, and the decoder is designed to effectively combine the features and predict the saliency map. During the past years, researchers have developed lots of feature fusion mechanisms and feature connection pathways for better representation. 
Liu~\textit{et al.}~\cite{liu2018picanet} proposed a hierarchical pixel-wise contextual attention network to learn the local and global context for each pixel. Zhao~\textit{et al.}~\cite{zhao2019pyramid} utilized channel attention for high-level representation and spatial attention for low-level feature maps to improve the detection performances. Our proposed two attention blocks share some similarities with~\cite{zhao2019pyramid} but are fundamentally different in three aspects. First, we do not distinguish channel attention or spatial attention for high-level or low-level features. Instead, we keep consistently using the combination of global and local attention for all the feature maps to be fused, regardless of its layer. Secondly, our attention blocks work in multiple scales and the fused representation would choose the information needed at a certain scale. Lastly, we have different architecture design and can take in two or three feature streams. 


\noindent \textbf{Edge-guided Models.}
In recent years, increasingly more approaches incorporate the edge/contour information to assist SOD task and improve the detection performances~\cite{zhang2017amulet,zhao2019pyramid,zhao2019egnet,qin2019basnet,su2019selectivity,liu2019simple,feng2019attentive,wu2019stacked,wei2020label}. 
Zhao~\textit{et al.}~\cite{zhao2019egnet} used edge label to supervise low-level feature maps to enable the network to have the capacity of modeling edge information. More recently, Wei~\textit{et al.}~\cite{wei2020label} proposed to explicitly decompose the ground truth label into the detail label that consists of pixels on the edge as well as pixels nearby the edge and the body label that concentrates on the pixels far from the edge. The two decoupled labels are used to supervise two branches in the first iteration, and a second iteration is still needed to train the fusion module for combing the results. Although we also decompose the original label, 
only the detail label is used to supervise intermediate results in our method. Another key difference is that our model is cascaded and can be trained end-to-end efficiently.   

\section{Methodology}
We start by introducing how the detail label is decomposed, then describe each part of the model in detail, and end with the loss function.

\begin{figure*} [!t] \small
	\centering
	\includegraphics[width=1\linewidth]{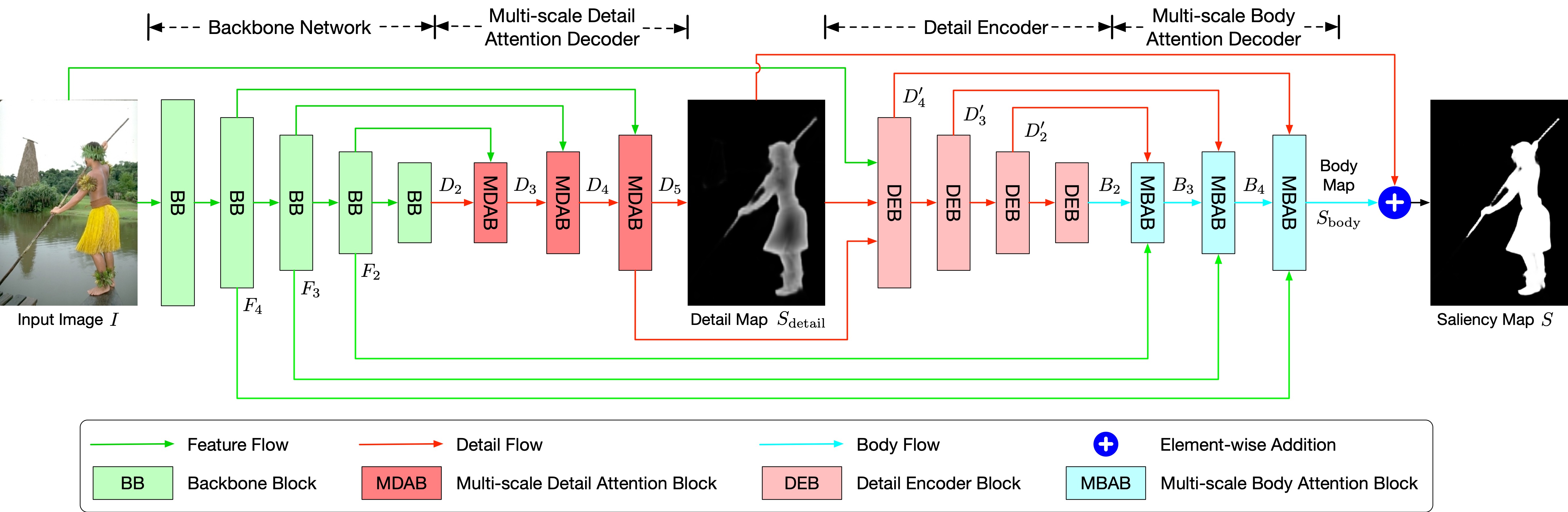}
	\caption{Overview of the proposed framework. The left sub-network consists of a backbone network and a proposed multi-scale detail attention decoder. The detail decoder takes in the feature flow transmitted from the backbone and outputs detail map. Subsequently, the detail encoder absorbs the input image, generated detail map, and the detail flow from the last MDAB. The fused feature is exploited by detail encoder to produce new detail flow for the body decoder. Then we feed the proposed multi-scale body attention decoder with detail and feature streams, where the contextual information is leveraged to predict the body map. The final saliency map is obtained by summing up the detail and body map.}
	\label{fig:framework}
\end{figure*}

\begin{figure}[!tbp] \small
    \centering
    \includegraphics[width=1\linewidth]{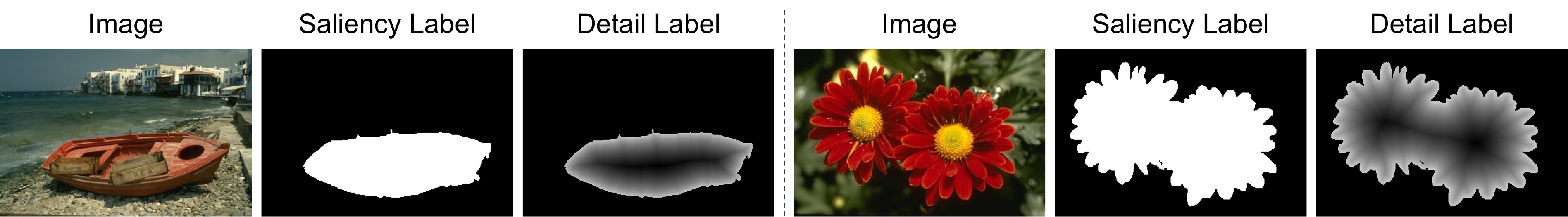}
    \caption{Some examples of decoupled detail labels. The pixels in decomposed label have larger values closer to the edge and smaller even zero value when far from the edge, which has a more balanced distribution than pure edge label.}
    \label{fig:detail_label_generation}
\end{figure}

\noindent \textbf{Detail Label Generation.}
As stated before, the pixels near the edge are hard to predict and prone to be misclassified. In the saliency detection task, the ground truth label is often binary and all the pixels in the salient regions have a unique value. Inspired by~\cite{wei2020label}, we explicitly decompose the detail label from the ground truth label and use it for our first sub-task detail modeling. More specifically, given the ground truth label $G$, we use the \emph{distance transformation} to convert the original label into the detail label in which each pixel in the original salient regions is defined by its minimum distance to the object boundary. This \emph{distance transformation} process can be described as:
\begin{equation}
  G_{\rm detail}(p,q)=\begin{cases}
  |G(p,q)-E(p,q)|, & G(p,q)=1,\\
  0, & G(p,q)=0,
  \end{cases}
\end{equation}
where $G_{\rm detail}$ represents the detail label, and $E(p,q)$ denotes the salient edge point that has the minimum Euclidean distance to saliency pixel $G(p,q)$. Fig.~\ref{fig:detail_label_generation} displays two examples of decomposed detail labels. After decoupling the detail label, it will be used to supervise the left sub-network in Fig.~\ref{fig:framework} to detect detail points close to the edge.

\begin{figure*} [!t] \small
	\centering
	\includegraphics[width=1\linewidth]{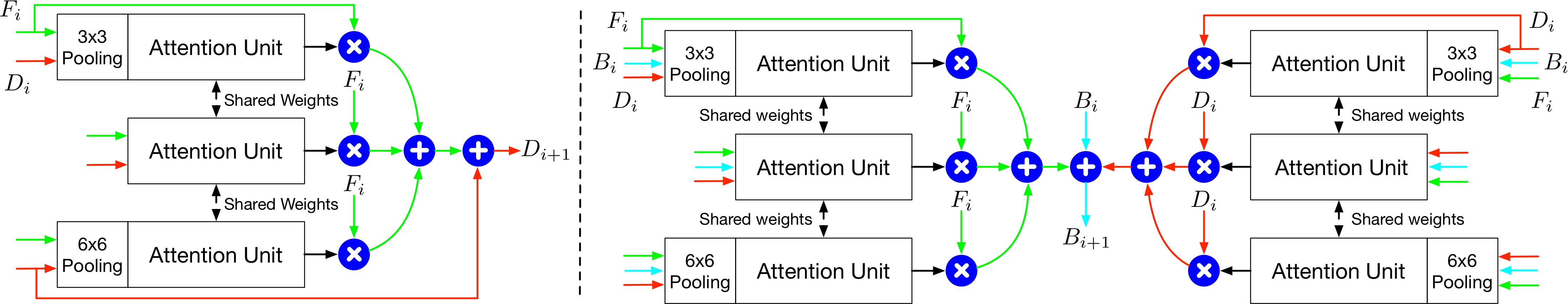}
	\caption{(\textbf{Left}) Workflow of our MDAB. The feature and detail streams are processed by three attention units to leverage contextual information and produce multi-scale attention for feature flow. Then the attentive feature flow is combined with detail flow for the generation of new detail flow. (\textbf{Right}) Pipeline of the proposed MBAB. We use three more attention units at different scales to generate attentive feature flow. The new body flow comes out of the summation of attentive detail flow, attentive feature flow, and original body flow.}
	\label{fig:mabb}
\end{figure*}

\noindent \textbf{Feature Extractor.}
Similar to existing models~\cite{wei2020f3net,zhao2019egnet,wei2020label}, we use ResNet-50~\cite{he2016deep} as our backbone network. We choose ResNet-50 as the backbone because it has the moderate model size and reasonable feature extraction power. Lager models can lead to better performances but will slow down the training and inference. The last fully connected layer and average pooling layer are removed, and we only keep the convolutional blocks. These blocks generate feature maps at five different scales $\{F_{i}|i{=}1,\dots,5\}$, where the resolution is down-sampled by two between subsequent blocks. As pointed out in~\cite{wu2019cascaded}, the representation $F_{5}$ at the shallowest layer contains too much coarse and redundant information. This increases computational burdens dramatically but brings little performance improvement. Hence, we abandon this layer and use only the rest finer features $\{F_{i}|i{=}1,\dots,4\}$. Two sets of the features will be passed to the detail decoder and body decoder respectively to assist their tasks.

\subsection{Multi-scale Detail Attention Modeling}

\begin{wrapfigure}{r}{0.5\linewidth}
	\centering
	\includegraphics[width=0.99\linewidth]{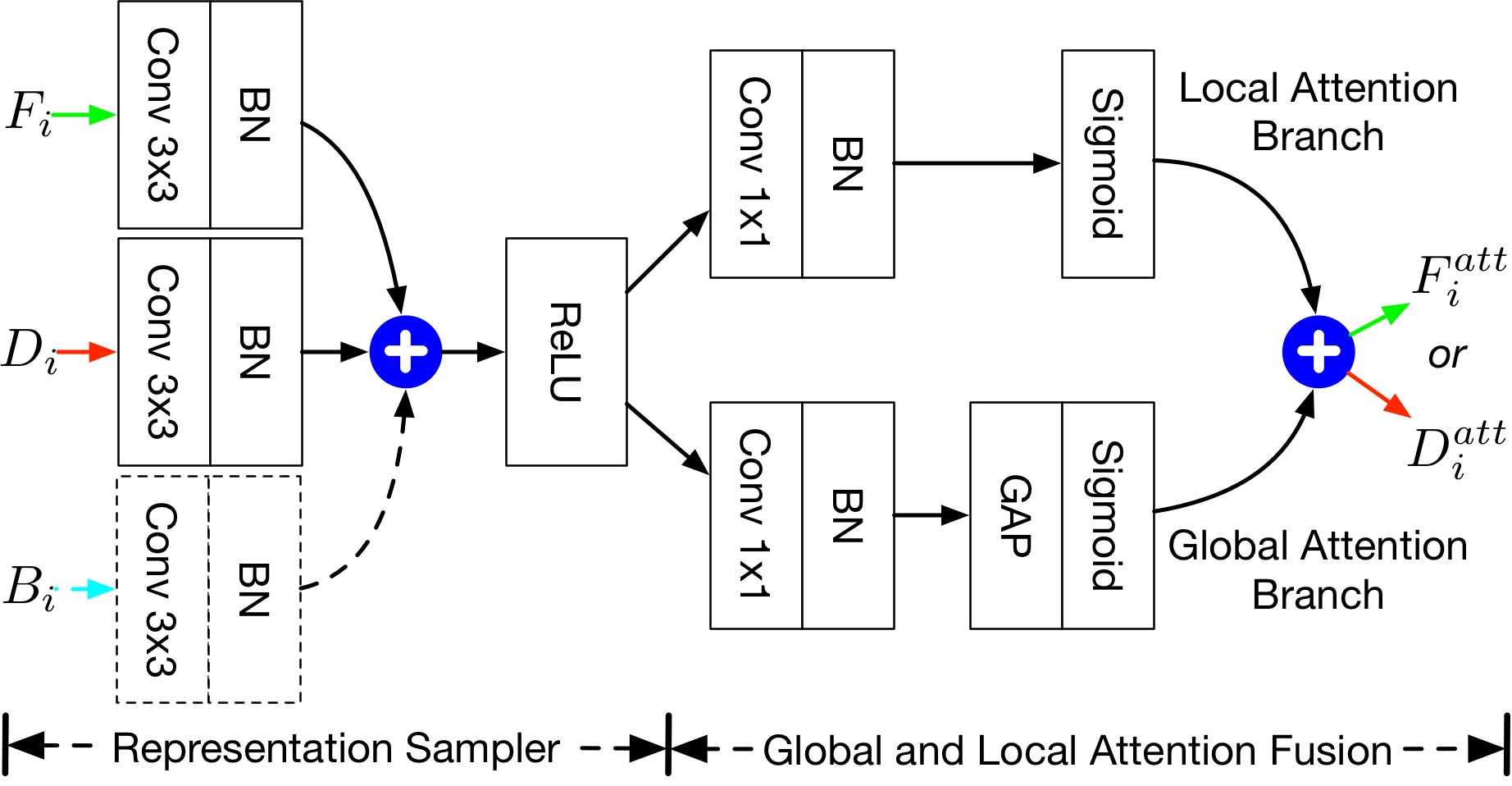}
	\caption{Architecture of attention unit. The attention unit works as a basic element for our MDAB and MBAB. Depending on the number of input streams, the representation sampler may have different number of convolution blocks. The unit is used to generate the combination of global and local attention for either the feature or detail flow.}
	\label{fig:unit}
\end{wrapfigure}

The detail decoder takes input of the image feature to fulfill the task of detail generation. It consists of three Multi-scale Detail Attention Block (MDAB). As shown in Fig.~\ref{fig:mabb} (\textit{left}), each MDAB absorbs both the detail flow from the block before and the feature flow from the backbone encoder at the corresponding scale, which can be denoted as:
\begin{equation}
    D_{i+1}={\rm MDAB}_{i}(D_{i},F_{i}),\  i=2,3,4
\end{equation}
where $F_{i}$ denotes the feature flow at the same level, $D_{i}$ stands for the current detail flow, and $D_{i+1}$ represents the new detail flow to be passed to the next block. At the last MDAB, we use $3{\times}3$ convolution layer and \textit{sigmoid} gate to extract detail map from the final detail flow. This operation can be denoted as, $S_{\rm detail}{=}\sigma({\rm Conv}(D_{5}))$,
where $\sigma(\cdot)$ denotes the \textit{sigmoid} gate, and ${\rm Conv}(\cdot)$ represents the convolution operation. The MDAB is mainly comprised of three attention units, where each one calculates the combination of local and global attention at one scale. The detailed architecture of the proposed attention unit is illustrated in Fig.~\ref{fig:unit}. Inside the attention unit, the detail flow and feature flow first go through one representation sampler to filter out useless noise and keep only the informative features:
\begin{equation}
    F^{att}_{i}={\rm ReLU}({\rm Conv}(D_{i})+{\rm Conv}(F_{i})),
\end{equation}
where the ReLU gate actively polishes the representation summed by detail flow and feature flow. Then the raw feature attention $F^{att}_{i}$ will be fed to two branches to obtain the corresponding local and global attention:
\begin{equation}
    F^{att}_{i}=\sigma({\rm Conv}(F^{att}_{i}))+\sigma({\rm GAP}({\rm Conv}(F^{att}_{i}))),
\end{equation}
where ${\rm GAP}$ denotes the global average pooling module. Assume the feature $F_{i}$ has the size of $B{\times} C{\times} H {\times} W$, the first term calculates the local spatial-wise attention of size $B{\times} 1 {\times} H {\times} W$, and the second term computes the global channel-wise attention of size $B{\times} C{\times} 1{\times} 1$. As can be seen from the left part of Fig.~\ref{fig:mabb}, the feature and detail flow pass three attention units after different pooling layers. We argue that the strategy of pooling with different kernels before the attention unit enables the feature to automatically search for useful information at a certain scale, which will benefit identifying salient objects of various sizes. Notice that the representation samplers of different attention units have shared weights, as we expect the sampler to have the function of filtering out the noise of fused flow, which should be robust against scale variation. In the end, we obtain new attentive detail flow that can identify both the object and its sharp boundaries:
\begin{equation}
    D_{i+1}={\rm Conv}(D_{i}+\sum_{t=1}^{3}(F_{i(t)}\odot F_{i(t)}^{att}))
\end{equation}
where $\odot$ represents element-wise multiplication, and $t$ denotes the number of scales used in MDAB.

\subsection{Multi-scale Body Attention Filling}

After the left detail decoder generates the detail map, the detail encoder in the right part of Fig.~\ref{fig:framework} will be fed with the input image, the detail flow, and the detail map to extract new detail flow $\{D^{'}_{i}|i{=}1,2,3,4\}$ for the task of body filling. Similar to the backbone network, the detail encoder will pass the detail flows to the body decoder via both normal path and short connections. Afterward, the body decoder takes in the image feature and detail feature to generate the body map. The body decoding task is completed by three subsequent Multi-scale Body Attention Block (MBAB). Each MBAB absorbs three streams, including the feature flow from the backbone network, the detail flow from the detail encoder, and the body flow from the previous block. This procedure can be represented as:
\begin{equation}
    B_{i+1}={\rm MBAB}(B_{i},F_{i},D^{'}_{i}),\ i=2,3,4
\end{equation}
where $B_{i}$ is the current body flow, and $B_{i+1}$ is the new body flow passing to the next block. After the three consecutive MBAB, we extract the body map and fill it with the detail map to generate the final saliency prediction $S{=}S_{\rm detail}{+}S_{\rm body}$. As we can see in Fig.~\ref{fig:mabb}(\textit{right}), our proposed MBAB resembles MDAB, with the main difference in three additional attention units dedicated for the detail flow. The detail and feature flow with multi-scale attention will be fused with body flow to create the new body stream. We can simply denote the MBAB workflow as:
\begin{equation}
    B_{i+1}={\rm Conv}(B_{i}+\sum_{t=1}^{3}(F_{i(t)}\odot F_{i(t)}^{att}+D^{'}_{i(t)}\odot D_{i(t)}^{'att})).
\end{equation}

\subsection{Hybrid Loss Function}

As we use both ground truth label and our decomposed detail label to train the network, we design different loss settings for the two tasks. For the detail output, we have the loss function defined as:
\begin{equation}
    l_{\rm detail}=l_{\rm CE}(S_{\rm detail},G_{\rm detail}) + l_{\rm SSIM}(S_{\rm detail},G_{\rm detail}),
\end{equation}
where the first term is the commonly used cross-entropy loss, and the second term is the structural similarity loss which enforces the detail decoder to focus on the edges. Structural similarity index measure (SSIM)~\cite{wang2003multiscale} was originally used to calculate the similarity of a pair of images by assessing the structural information. Motivated by~\cite{qin2019basnet}, we also integrate this loss to let the detail output learn the structural information of the image for keeping the precise edges. This loss is calculated as:
\begin{equation}
    l_{\rm SSIM}=1-\frac{(2\mu_{x}\mu_{y}+C_{1})(2\sigma_{xy}+C_{2})}{(\mu_{x}^2+\mu_{y}^{2}+C_{1})(\sigma_{x}^{2}+\sigma_{y}^{2}+C_{2})},
\end{equation}
where $\mu_{x}$, $\mu_{y}$ and $\sigma_{x}$, $\sigma_{y}$ are the mean and standard deviation of the image. $C_{1}$ and $C_{2}$ are small positive constants, and we set them as $0.01^2$ and $0.03^2$ to avoid dividing zero. As for the body mask prediction, we have the following loss configuration:
\begin{equation}
    l_{\rm body}=l_{\rm CE}(S,G) + l_{\rm IoU}(S,G)+l_{\rm F}(S,G),
    \label{bodyloss}
\end{equation}
where the second term is the IoU loss, which can help the body decoder quickly attend to the main body of the object as adopted in~\cite{rahman2016optimizing,mattyus2017deeproadmapper,qin2019basnet}. This loss can be computed as:
\begin{equation}
    l_{\rm IoU}=1-\frac{\sum_{i=1}^{H}\sum_{j=1}^{W}S(i,j)G(i,j)}{\sum_{i=1}^{H}\sum_{j=1}^{W}(S(i,j)+G(i,j)-S(i,j)G(i,j))}.
\end{equation}
The third term in Eq.~\eqref{bodyloss} is the so-called F-loss~\cite{zhao2019optimizing}. It is proposed to directly optimize the metric F-measure as defined by: $l_{\rm F}{=}1{-}F(S,G)$.
Here we expect that adopting this loss can balance the detail and body map to complement the information of each other by pushing their fused mask to achieve a high F-measure score.
In total, our model is trained end-to-end using the hybrid loss function:
\begin{equation}
    l=\frac{1}{2}(l_{\rm detail}+l_{\rm body}).
\end{equation}

\section{Experiments}
\label{sec:ex}

\noindent \textbf{Datasets.}
Following \cite{wu2019cascaded,wu2019stacked,wei2020f3net,wei2020label}, we conduct extensive experiments on six widely used benchmark datasets to evaluate the effectiveness of the proposed method, \emph{i.e.}, ECSSD~\cite{yan2013hierarchical}, PASCAL-S~\cite{li2014secrets}, DUT-OMRON~\cite{yang2013saliency}, HKU-IS~\cite{li2016visual}, THUR15K~\cite{cheng2014salientshape}, and DUTS~\cite{wang2017learning}. 

Specifically, ECSSD~\cite{yan2013hierarchical} contains 1,000 structurally complex natural images. PASCAL-S~\cite{li2014secrets} consists of 850 images with cluttered backgrounds chosen from the validation set of the PASCAL-VOC segmentation dataset~\cite{everingham2010pascal}. DUT-OMRON~\cite{yang2013saliency} has 5,168 images with high content variety. HKU-IS~\cite{li2016visual} has 4,447 images containing mostly  multiple disconnected objects. THUR15K~\cite{cheng2014salientshape} consists of 6,232 diverse and heterogeneous images categorized into several groups. Among these datasets, DUTS \cite{wang2017learning} is currently the largest saliency detection dataset consisting of two subsets: DUTS-TR contains 10,553 images for training and DUTS-TE has 5,019 images for testing.

\noindent \textbf{Implementation Details.}
\label{sec:3}
In line with most existing methods~\cite{qin2019basnet,wei2020f3net,zhao2020suppress,pang2020multi,wei2020label}, we use the DUTS-TR dataset for training and the rest of the datasets as the test set for evaluation. ResNet-50~\cite{he2016deep} classifier pre-trained on ImageNet~\cite{deng2009imagenet} is used as backbone to initialize the model, and the other parameters are randomly initialized. Our network is trained end-to-end for 50 epochs with a mini-batch size of 32 by stochastic gradient descent (SGD). The momentum and weight decay are set to 0.9 and 0.0005, respectively. We set the maximum
learning rate to 0.005 for the ResNet-50 backbone and 0.05 for the other parts. Warm-up and linear decay strategies are also used. 

During training, we use random horizontal flip, random crop, and multi-scale input images for data augmentation. The images are resized to the resolution of $352{\times}352$ during testing and fed into the network to generate the saliency prediction without any post-processing step. Resizing with bilinear interpolation is consistently used throughout all the experiments. The proposed model achieves the inference time of 20 FPS on a single Quadro RTX 6000 GPU.

\noindent \textbf{Evaluation Metrics.} We use four widely used metrics to evaluate the proposed method, \emph{i.e.}, Mean Absolute Error ($MAE$)~\cite{perazzi2012saliency}, mean F-measure (m $F_{\beta}$)~\cite{achanta2009frequency}, weighted F-measure ($F_{\beta}^{\omega}$)~\cite{margolin2014evaluate}, and precision-recall curve. pecifically, $MAE$~\cite{perazzi2012saliency} calculates the absolute per-pixel difference between the saliency prediction and its ground truth: 
\begin{equation}
    MAE=\frac{1}{W\times H}\sum_{i=1}^{W}\sum_{j=1}^{H}|{S(i,j)-G(i,j)}|,
\end{equation}
where $W$ and $H$ is the width and height of the mask, $S$ denotes the predicted saliency map, and $G$ represents the ground truth. As the most fundamental and direct measure, $MAE$ has been widely applied to evaluate the quality of saliency map~\cite{qin2019basnet,wei2020f3net,zhao2020suppress,pang2020multi,wei2020label}. The generated saliency map $S$ is first converted into a  binary map using a threshold and is compared with ground truth $G$ to compute the \textit{precision} and \textit{recall} score:
\begin{equation}
    \begin{gathered}
    precision=\frac{|S\cap G|}{S},\ recall=\frac{|S\cap G|}{G}
    \end{gathered}
\end{equation}
The precision-recall curve is plotted by varying the binarized thresholds from 0 to 255 to obtain a sequence of precision-recalled pairs. The larger the area under the PR curve, the better the performance of the model. F-measure $F_{\beta}$ and its weighted variant $F_{\beta}^{\omega}$ are used to jointly assess the saliency prediction by taking both \textit{precision} and \textit{recall} into consideration. The basic F-measure can be formulated as:
\begin{equation}
    F_{\beta}=\frac{(1+\beta^2) { precision}\times { recall}}{\beta^{2} { precision} + { recall}},
\end{equation}
where $\beta$ is the relative weight to control the importance of \textit{precision} and \textit{recall}. $\beta$ is usually set to $0.3$ to give a larger weight to \textit{precision} as suggested in~\cite{achanta2009frequency}. Mean F-measure (m $F_{\beta}$) is computed by taking the mean value of F-measure from the PR curve. Weighted F-measure~\cite{margolin2014evaluate} is an intuitive generalization of F-measure for non-binary maps, which is defined as:
\begin{equation}
    F_{\beta}^{\omega}=\frac{(1+\beta^2){ precision}^{\omega}\times { recall}^{\omega}}{\beta^{2}{ precision}^{\omega}+{ recall}^{\omega}}
\end{equation}
The basic quantities $precision$ and $recall$ are extended to non-binary values and assigned different weights to different errors according to location and neighborhood information. Except for these four measures, Max F-measure that selects the maximum value of F-measure from the PR curve, E-measure~\cite{fan2018enhanced}, and S-measure~\cite{fan2017structure} are also used in the literature. S-measure is proposed to compute the region-aware and object-aware similarities, and E-measure is designed to combine local pixel values with image-level mean values for joint assessment.

\begin{figure*}[t] \small
    \centering
    \includegraphics[width=1\linewidth]{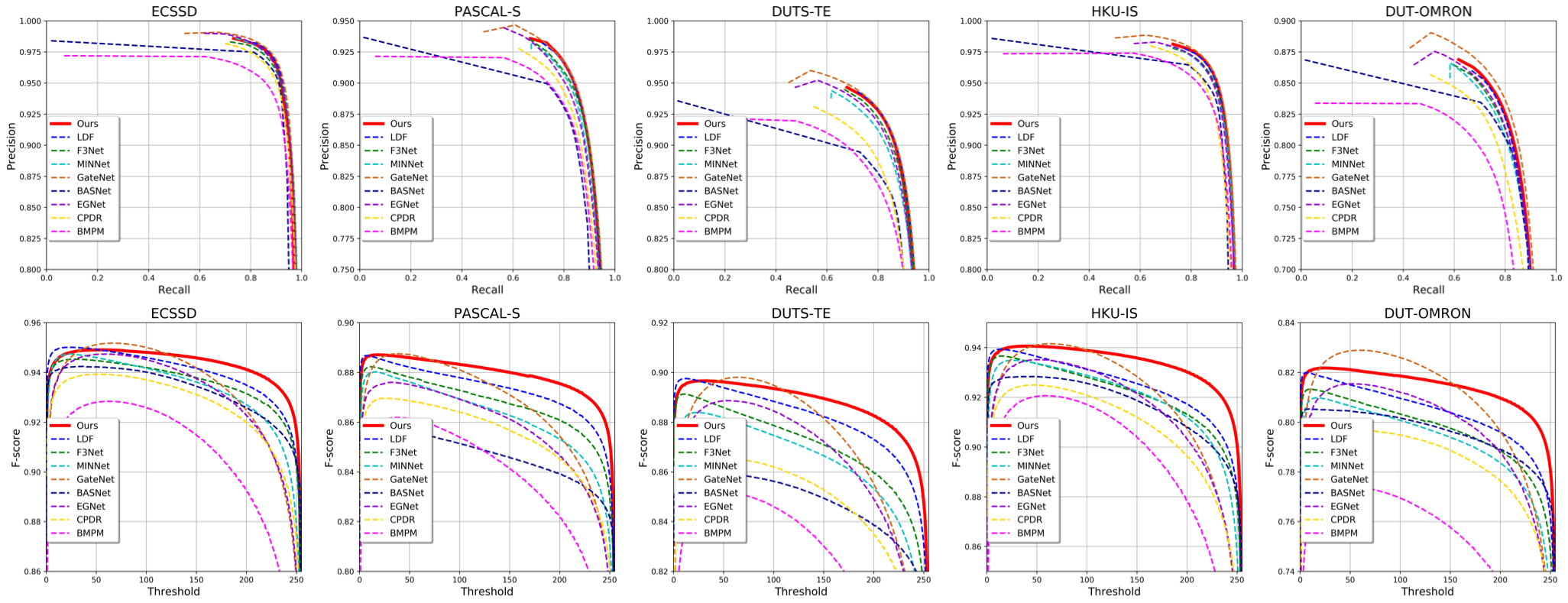}
    \caption{The precision-recall curve (\textbf{first row}) and F-measure versus different thresholds (\textbf{second row}) of all the methods.}
    \label{fig:PR_F_curve}
\end{figure*}

\begin{table*}[!t] \small
    \centering
    \caption{Quantitative results compared with state-of-the-art methods on six datasets. `--' means the results can not be obtained. For all metrics except for $MAE$, higher is better. 
    The best three results are highlighted in \textbf{\textcolor{red}{red}}, \textcolor{blue}{blue}, and \textcolor{green}{green} respectively.}
    \resizebox{0.99\linewidth}{!}{
    \begin{tabular}{r|ccc|ccc|ccc|ccc|ccc|ccc} \toprule
         \multirow{2}*{Method$_{\rm year}$}& \multicolumn{3}{c|}{ECSSD (\#1,000)}& \multicolumn{3}{c|}{DUTS-TE (\#5,019)} & \multicolumn{3}{c|}{DUT-OMRON (\#5,168)} & \multicolumn{3}{c|}{PASCAL-S (\#850)} & \multicolumn{3}{c|}{HKU-IS (\#4,447)} & \multicolumn{3}{c}{THUR15K (\#6,232)}    \\
         \cline{2-19} 
         &$MAE$&m $F_{\beta}$&$F_{\beta}^{\omega}$
         &$MAE$&m $F_{\beta}$&$F_{\beta}^{\omega}$
         &$MAE$&m $F_{\beta}$&$F_{\beta}^{\omega}$
         &$MAE$&m $F_{\beta}$&$F_{\beta}^{\omega}$
         &$MAE$&m $F_{\beta}$&$F_{\beta}^{\omega}$ 
         &$MAE$&m $F_{\beta}$&$F_{\beta}^{\omega}$ \\ \midrule
         BMPM$_{2018}$~\cite{zhang2018bi} &0.045&0.868&0.871 
         &0.049&0.745&0.761 
         &0.064&0.692&0.681 
         &0.076&0.769&0.782 
         &0.039&0.871&0.859 
         &0.079&0.704&-- 
         \\
         \hline
         CPD-R$_{2019}$~\cite{wu2019cascaded} &0.037&0.917&0.898 
         &0.043&0.805&0.795 
         &0.056&0.747&0.719 
         &0.074&0.829&0.800 
         &0.034&0.891&0.875 
         &0.068&0.738&0.730 
         \\
         \hline
         EGNet-R$_{2019}$~\cite{zhao2019egnet} &0.037&0.920&0.903 
         &0.039&0.815&0.816 
         &\textcolor{green}{0.053}&0.756&0.738 
         &0.075&0.831&0.807 
         &0.031&0.901&0.887 
         &0.067&0.739&0.733 
         \\
         \hline
         BANet$_{2019}$~\cite{su2019selectivity} &\textcolor{green}{0.035}&0.923&0.908 
         &0.040&0.815&0.811 
         &0.059&0.746&0.736 
         &0.070&0.838&0.817 
         &0.032&0.899&0.887 
         &0.068&0.741&-- 
         \\
         \hline
         BASNet$_{2019}$~\cite{qin2019basnet} &0.037&0.880&0.904 
         &0.048&0.791&0.803 
         &0.056&0.756&\textcolor{green}{0.751} 
         &0.079&0.777&0.797 
         &0.032&0.895&0.889 
         &0.073&0.733&0.721 
         \\
         \hline
         SCRN$_{2019}$~\cite{wu2019stacked} &0.037&0.918&0.899  
         &0.040&0.809&0.803 
         &0.056&0.746&0.720 
         &\textcolor{green}{0.065}&0.839&0.816 
         &0.033&0.897&0.878 
         &0.066&0.741&0.734 
         \\
         \hline
         F3Net$_{2020}$~\cite{wei2020f3net} &\textbf{\textcolor{red}{0.033}}&\textcolor{green}{0.925}&\textcolor{blue}{0.912} 
         &\textcolor{green}{0.035}&0.791&\textcolor{green}{0.835} 
         &\textcolor{green}{0.053}&\textcolor{green}{0.766}&0.747 
         &\textcolor{blue}{0.064}&\textcolor{green}{0.844}&\textcolor{green}{0.823} 
         &\textcolor{blue}{0.028}&\textcolor{green}{0.910}&\textcolor{green}{0.900} 
         &\textcolor{green}{0.065}&\textcolor{blue}{0.756}&\textcolor{green}{0.744} 
         \\
         \hline
         GateNet$_{2020}$~\cite{zhao2020suppress} &\textcolor{green}{0.035}&0.917&0.906 
         &\textcolor{green}{0.035}&0.816&0.828 
         &\textcolor{blue}{0.051}&0.761&0.749 
         &\textcolor{green}{0.065}&0.827&0.821 
         &\textcolor{green}{0.029}&0.903&0.893 
         &--&--&-- 
         \\
         \hline
         MINet$_{2020}$~\cite{pang2020multi} 
         &\textbf{\textcolor{red}{0.033}}&0.924&\textcolor{green}{0.911} 
         &0.037&\textcolor{green}{0.828}&0.825 
         &0.055&0.756&0.738 
         &\textcolor{blue}{0.064}&0.842&0.821 
         &\textcolor{blue}{0.028}&0.908&0.899 
         &--&--&-- 
         \\
         \hline
         LDF$_{2020}$~\cite{wei2020label}  &\textcolor{blue}{0.034}&\textcolor{blue}{0.930}&\textbf{\textcolor{red}{0.915}} 
         &\textcolor{blue}{0.034}&\textcolor{blue}{0.855}&\textcolor{blue}{0.845} 
         &\textcolor{blue}{0.051}&\textcolor{blue}{0.773}&\textcolor{blue}{0.752} 
         &\textbf{\textcolor{red}{0.062}}&\textcolor{blue}{0.853}&\textcolor{blue}{0.828} 
         &\textbf{\textcolor{red}{0.027}}&\textcolor{blue}{0.914}&\textcolor{blue}{0.904} 
         &\textcolor{blue}{0.064}&\textcolor{blue}{0.763}&\textcolor{blue}{0.752} 
         \\
         \hline
         Ours &\textcolor{green}{0.035}&\textbf{\textcolor{red}{0.931}}&\textcolor{green}{0.911} 
         &\textbf{\textcolor{red}{0.033}}&\textbf{\textcolor{red}{0.863}}&\textbf{\textcolor{red}{0.847}} 
         &\textbf{\textcolor{red}{0.048}}&\textbf{\textcolor{red}{0.785}}&\textbf{\textcolor{red}{0.758}} 
         &\textbf{\textcolor{red}{0.062}}&\textbf{\textcolor{red}{0.855}}&\textbf{\textcolor{red}{0.829}} 
         &\textbf{\textcolor{red}{0.027}}&\textbf{\textcolor{red}{0.924}}&\textbf{\textcolor{red}{0.907}}
         &\textbf{\textcolor{red}{0.062}}&\textbf{\textcolor{red}{0.769}}&\textbf{\textcolor{red}{0.755}} 
    \\ \bottomrule
    \end{tabular}
    }
    \label{tab:sub_table}
\end{table*}

\subsection{State-of-the-Art Comparisons}
\noindent \textbf{Quantitative Evaluation.}
We demonstrate the efficacy of our model by comparing with other 10 most recent state-of-the-art models, including BMPM~\cite{zhang2018bi}, CPD~\cite{wu2019cascaded}, EGNet~\cite{zhao2019egnet}, BANet~\cite{su2019selectivity}, BASNet~\cite{qin2019basnet}, SCRN~\cite{wu2019stacked}, F3Net~\cite{wei2020f3net}, GateNet~\cite{zhao2020suppress}, MinNet~\cite{pang2020multi}, and LDF~\cite{wei2020label}. To assure comparison fairness, the saliency maps are either provided by the authors or generated using officially released pre-trained models. 
Table~\ref{tab:sub_table} displays the performances of aforementioned methods on six datasets. Our method consistently outperforms other models and achieves the best performances across six datasets, refreshing the leaderboard and setting the new baseline. In particular, we have significantly improved the best F-score ($F_{\beta}$) over all datasets, with 1.5\% increase on DUT-OMRON, 1.1\% on HKU-IS, and 0.9\% on DUTS-TE. It is also worth mentioning that our method surpasses others by larger margin on large datasets, while the difference on small-scale datasets ($<$1,000 images) is less obvious. Due to the limited number of images, small datasets may not well reflect the actual performance of a model.

Fig.~\ref{fig:PR_F_curve} shows the precision-recall curve ($1_{st}$ row) and the F-measure curve ($2_{nd}$ row) of all the methods. Our PR curve consistently lies above other methods and achieves best performances on ECSSD, PASCAL-S, DUTS-TE, and HKU-IS, and has very competitive results on DUT-OMRON. Moreover, our PR curve is significantly shorter than other methods and has larger recall value ranges, which indicates that our method has less \textit{false negative} predictions in the saliency maps. Across all datasets, our F-measure curve has the flattest slope and largest area under the curve, demonstrating that our generated saliency maps present good quality against varying thresholds. 

Notice that on ECSSD and DUT-OMRON the GateNet~\cite{zhao2020suppress} seem to have the highest F-score at certain thresholds but the score drops very quickly when the threshold is increased. This implies that their saliency maps have many non-binary predictions (\emph{i.e.,} numerical values on (0,1)). Only when an appropriate threshold is carefully chosen, their method could have reasonable performances. 

\begin{figure*}[t] \small
    \centering
    \includegraphics[width=1\linewidth]{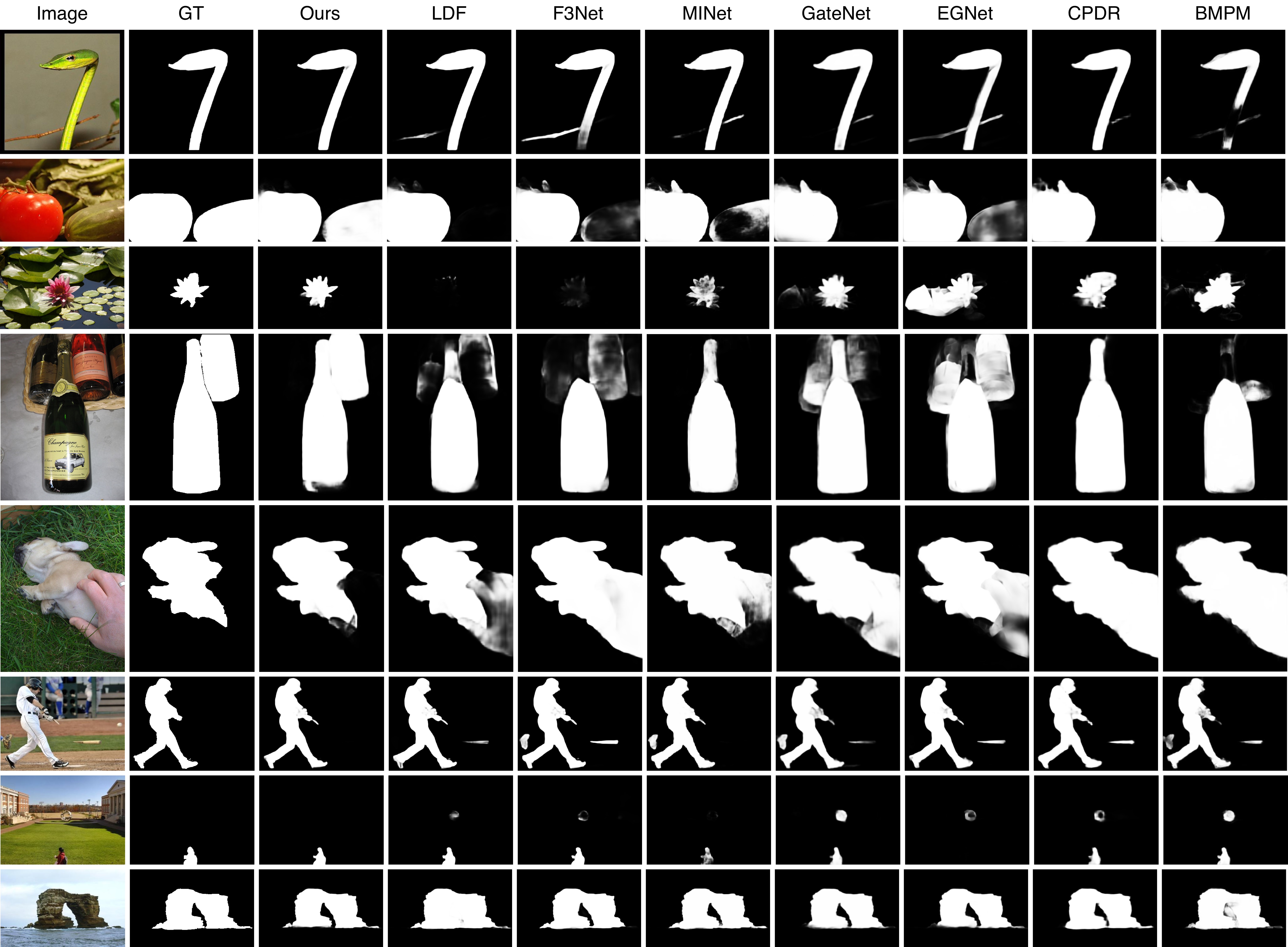}
    \caption{Visual comparison of our method with other state-of-the-art methods in different challenging scenarios. Our method can well distinguish salient objects and suppress background noise, giving better visual appeal than others.}
    \label{fig:visual_comparison}
\end{figure*}

\noindent \textbf{Qualitative Evaluation.}
Some representative visual examples are shown in Fig.~\ref{fig:visual_comparison}. We select images from some challenging scenarios, including low color contrast ($1_{st}$ row), high inter-object contrast ($2_{nd}$ row), low contrast near object boundary ($3_{rd}$ row), multiple objects with low background contrast ($4_{th}$ row), partly occluded object ($5_{th}$ row), object in cluttered backgrounds ($6_{th}$ row), small object near image border ($7_{th}$ row), and object with irregular and complex edges (last row). It can be seen that our method well suppresses background noise and accurately segments the salient objects of various sizes with coherent details.


\subsection{Ablation Study}

\noindent \textbf{Baseline Models.}
To investigate the effect of each proposed module, we conduct ablation studies on several baselines to validate the effectiveness of each proposed component: 1) B1 uses ResNet backbone and a decoder network for direct saliency prediction; 2) B2 first generates intermediate body map and then produces detail map to refine the boundary; 3) B3 models the detail first and then fills the body map into the detail mask; 4) B6 uses structural similarity loss to enforce the detail decoder to learn the structural information; 5) B7 adopts IoU loss to help body decoder quickly attend to body and F-loss to balance the body and detail information. 6) B4 additionally employs MDAB for better detail modeling; 7) B5 applies MBAB to fuse the detail, feature for body generation; 
Table~\ref{tab:abla} shows the results of the ablation study on THUR15K. As we deploy more proposed modules, the performance gains step-wise improvement, demonstrating the effectiveness of each proposed module. We then analyze the impact of each module in the following paragraphs.

\begin{figure*}[!t] \small
    \centering
    \includegraphics[width=1\linewidth]{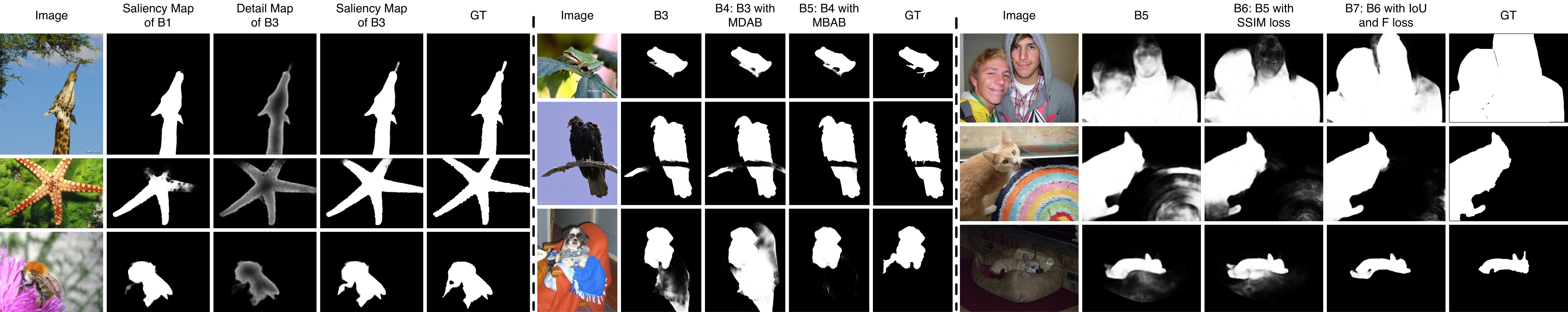}
    \caption{(\textbf{Left}) Visual illustration of detail modeling. The first stage of B3 aims at identifying the informative details of the object, which benefits the downstream body filling task for more accurate saliency map generation. (\textbf{Middle}) Effect of multi-scale attention blocks. Equipped with the two proposed blocks, the network learns to attend to the crucial regions and remove noise to refine the mask. (\textbf{Right}) Visual effect of the hybrid loss function. $l_{\rm SSIM}$ improves the mask by enriching the representation with the structural information, while $l_{\rm IoU}$ and $l_{\rm F}$ helps the network to concentrate on the body and complements the body and detail map by ensuring the fused mask has a high F-score, respectively.}
    \label{fig:ablatioin}
\end{figure*}

\noindent \textbf{Effect of Detail Modeling.} 
We evaluate the impact of detail modeling by comparing the baseline B3 that captures detail first to baseline B1 that directly outputs the saliency map. As can be seen from Table~\ref{tab:abla}, detail modeling brings about 1.5\% increase in terms of both $F_{\beta}$ and $F_{\beta}^{\omega}$. Fig.~\ref{fig:ablatioin} (\emph{left}) illustrates the visual impact of detail modeling. We can see that the detail map can first identify the informative edges of the object and thus help the body filling part to generate saliency map with more accurate boundaries. 

\begin{table}[!t] \small
	\centering
	\caption{\small Ablation studies on {\small THUR15K}.
	}
		\resizebox{0.6\linewidth}{!}{%
	\begin{tabular}{clccc} \toprule
		& Setting  &  $MAE \downarrow$ & m $F_{\beta} \uparrow$ & $F_{\beta}^{\omega} \uparrow$  \\ \midrule	
		B1 & Baseline & 0.071 & 0.726 & 0.721 \\
		B2 & B1 + Body Map $\rightarrow$ Detail Map & 0.070 & 0.732 &0.727  \\
		B3 & B1 + Detail Map $\rightarrow$ Body Map & 0.068 & 0.737 &0.732\\ 
		B4 & B3 + $l_{\rm SSIM}$ on Detail Map &0.066  &0.749  &0.744 \\
		B5 & B4+ $l_{\rm IoU}$ and $l_{\rm F}$ on Body Map &0.065  &0.753 &0.750\\ 
	
		B6 & B5 + MDAB  &0.064  &0.760 &0.752     \\
		B7 & B6 + MBAB &\textbf{0.062}  &\textbf{0.769} &\textbf{0.755}             \\
		\bottomrule
	\end{tabular}
	}
	\label{tab:abla}
\end{table}


\noindent \textbf{Impact of Generation Order.}
We design an interesting baseline B2 that generates intermediate body map to investigate the impact of generation order, namely taking the strategy ``easier first'' to let the network focus on the easier body map task and then refine the boundary, or the strategy ``harder first'' to make the model learn the harder detail first and then fill in the body. From Table~\ref{tab:abla}, we can see that B3 surpasses B2, proving the effectiveness of ``harder first'' strategy. The reason behind may be that the neural network naturally focuses on the low-level information like the edges first then gradually shift to high-level semantics.

\noindent \textbf{Effect of Multi-scale Attention Block.}
Based on B3 with detail modeling, we demonstrate the effect of our proposed MDAB and MBAB by setting baseline B4 and B5. Table~\ref{tab:abla} tells that the successive deployment of the two attention blocks improves the baseline by 1\% in $F_{\beta}$ and $F_{\beta}^{\omega}$. As we can see from the visual illustration shown in Fig.~\ref{fig:ablatioin} (\emph{middle}), the two attention blocks enforce the model to concentrate on the salient object and refine the mask by removing the misclassified background region.

\noindent \textbf{Effect of Hybrid Loss Function.}
To validate the effect of the hybrid loss function, we conduct a set of experiments over different loss configurations on our model. The evaluation results in Table~\ref{tab:abla} show that the combination of structural similarity loss on the detail map and the IoU and F loss on the body map works best. Fig.~\ref{fig:ablatioin} (\emph{right}) displays some visual examples to further demonstrate this. We can observe that $l_{\rm SSIM}$ can effectively complement the structural information and results in saliency maps with sharp and clear boundaries. On the other hand, $l_{\rm IoU}$ and $l_{\rm F}$ can ensure that the network focuses on the salient object and well combine the two maps to achieve a high F-score. 

\begin{figure*}[!t] \small
    \centering
    \includegraphics[width=1\linewidth]{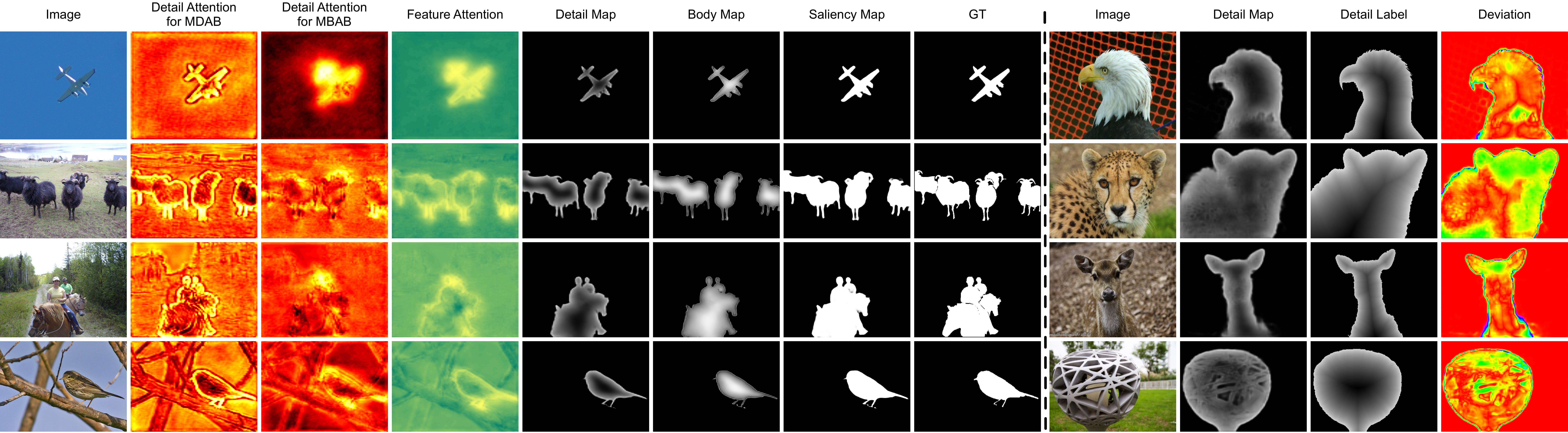}
    \caption{(\textbf{Left}) Visual examples of the attention and output maps. (\textbf{Right}) Visualization of the deviation between predicted detail map and decomposed detail label. The detail map may also assign values to features that can characterize the object.}
    \label{fig:attention_visualization}
\end{figure*}

\noindent \textbf{Visualizing Attention, Detail, and Body Maps.}
To understand how the multi-scale attention works, we visualize the local spatial-wise attention maps in the last MBAB and MDAB, the detail and body map, and the fused saliency map in Fig.~\ref{fig:attention_visualization} (\emph{left}). As can be seen, the first detail attention map for detail decoder highlights the edges of the object, pushing the network to concentrate on the boundary. The second detail attention for body decoder mainly emphasizes some crucial details and reveals regions that the previous attention map may neglect. The feature attention map pays attention to the important features of the image and encourages the network to segment more accurate saliency maps. 

\begin{figure} [t] \small
    \centering
    \includegraphics[width=1\linewidth]{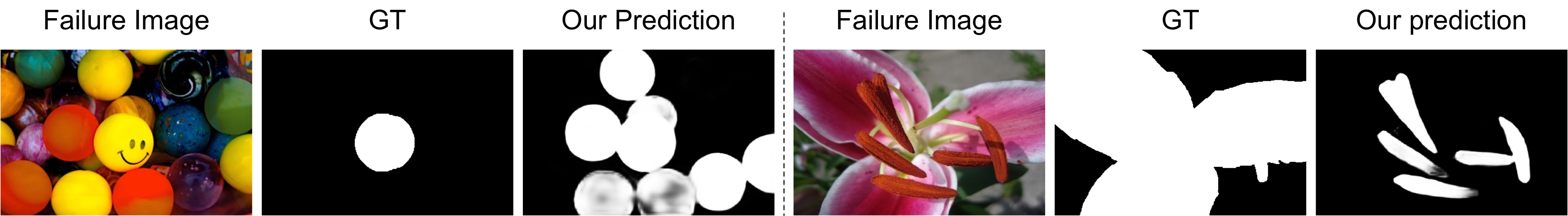}
    \caption{Examples of two failure cases of our model.}
    \label{fig:failure_cases}
\end{figure}

\noindent \textbf{Deviation of Detail Map.}
To study the concrete effect of detail label, we measure the deviation between generated detail map and explicitly decoupled detail label and present some examples in Fig.~\ref{fig:attention_visualization} (\emph{right}). We can see that the generated detail maps do not necessarily follow the exact distribution of the decoupled detail label that only assigns larger values to pixels nearby edge. Instead, the produced detail map may also assign values to crucial features that characterize the object (\emph{e.g.}, the neck of the eagle). We expect the decoupled detail map plays the role that leads the detail map to distinguish crucial pixels based on the detail label.


\noindent \textbf{Failure Cases.}
Fig.~\ref{fig:failure_cases} presents two examples of failure cases on ECSSD where our model has the narrowest margin over other methods. The left example is an image that has a ``smiling'' ball in the center. The ground truth displays only the ``smiling'' ball, whereas our map predicts all the balls of high color contrast. The reason may be that our model focuses too much on the background contrast but fails to handle the inter-object contrast in this specific instance. The right example is a flower where the ground truth presents the whole flower but ours only predict its stamen. The stamen does naturally pops out of the image but is only a sub-object of the flower. We think it is because there is a lack of image-level class label supervision to let the network learns the category of an object. 

\section{Conclusion}
We propose a novel end-to-end SOD framework that disentangles the original task into cascaded detail modeling and body filling. This framework can effectively reduce the difficulty of direct saliency detection. Moreover, we propose two multi-scale attention blocks that target feature fusion and help the network to generate more accurate detail and body maps. Extensive experiments have demonstrated that our method achieves state-of-the-art performances on different metrics across six datasets. 
\begin{acks}
This work has been supported by the EU H2020 AI4Media (No. 951911).
\end{acks}


\appendix
\section{Appendix}

This document  provides additional experimental results of our saliency detection model. 
First, we show additional visual examples to compare the proposed method with several state-of-the-art methods (Sec.~\ref{sec:4}), \emph{i.e.}, LDF~\cite{wei2020label}, F3Net~\cite{wei2020f3net}, MINet~\cite{pang2020multi}, GateNet~\cite{zhao2020suppress}, CPDR~\cite{wu2019cascaded}, and BMPM~\cite{zhang2018bi}. The visualization results of each ablation model is also displayed (Sec.~\ref{sec:5}).

\begin{figure*}
    \centering
    \includegraphics[width=1.0\linewidth]{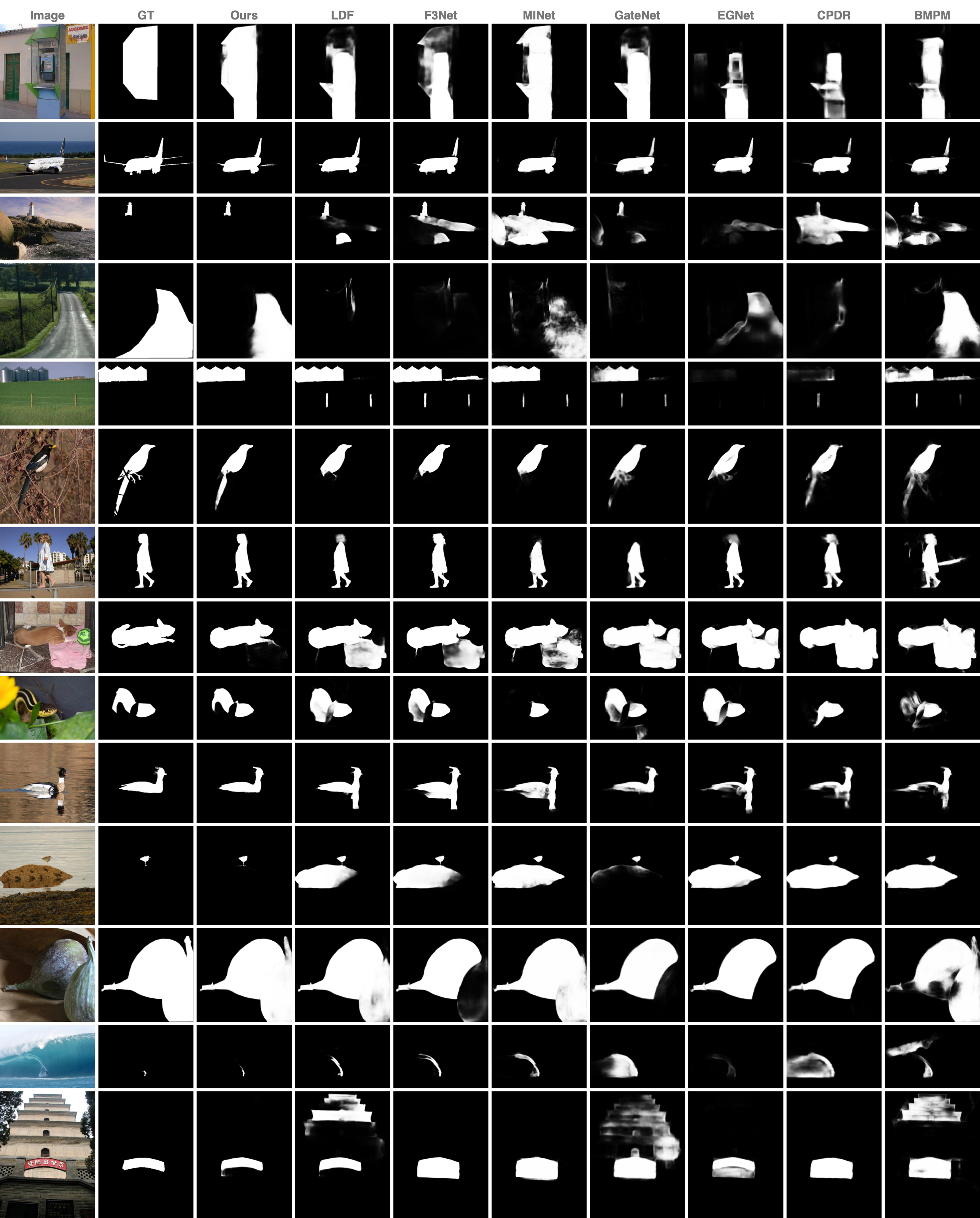}
    \caption{Visualization of our model against other state-of-the-art methods, \emph{i.e.}, LDF~\cite{wei2020label}, F3Net~\cite{wei2020f3net}, MINet~\cite{pang2020multi}, GateNet~\cite{zhao2020suppress}, CPDR~\cite{wu2019cascaded}, and BMPM~\cite{zhang2018bi}. }
    \label{fig:visual_comparison_appendix}
\end{figure*}

\begin{figure*}
    \centering
    \includegraphics[width=1.0\linewidth]{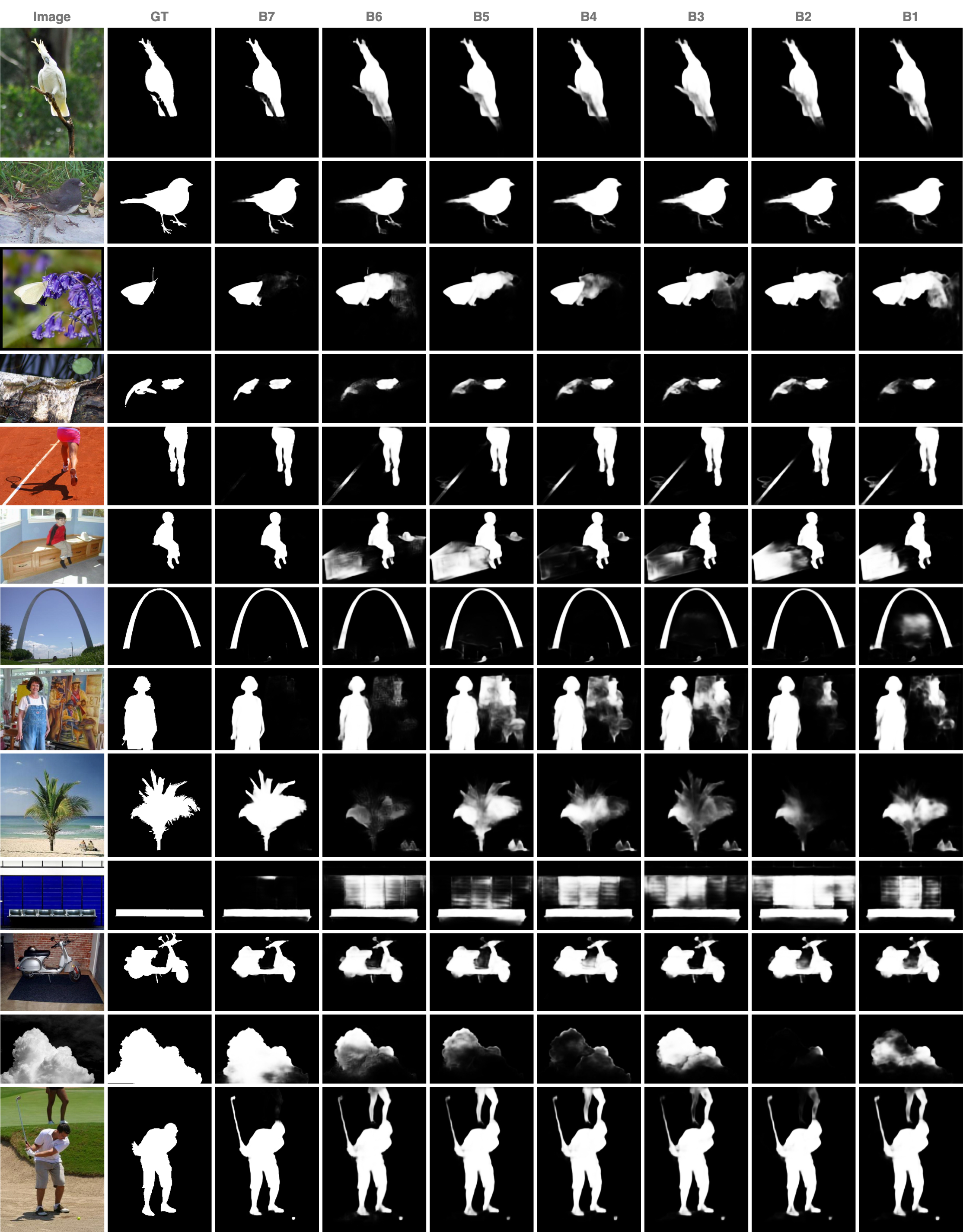}
    \caption{Visual examples of all the baseline models used in the ablation studies. }
    \label{fig:ablation_appendix}
\end{figure*}

\subsection{Visual Comparison against State-of-the-Art}
\label{sec:4}
We show some example images of our model and existing leading methods in Fig.~\ref{fig:visual_comparison_appendix}. Thanks to the proposed detail modeling and multi-scale attention mechanism, our method can well distinguish single or multiple salient objects of various sizes from cluttered and complex backgrounds.

\subsection{Visualization of Ablation Models}
\label{sec:5}
To investigate how the predicted saliency maps manifest as more proposed modules are employed, we show some example images of all the ablation models in Fig.~\ref{fig:ablation_appendix}. As can be seen, the generated masks evolve and present step-wise visual improvement. Each deployed module brings about either refinement on details or elimination of misclassified regions, leading to better saliency masks.

\bibliographystyle{ACM-Reference-Format}
\bibliography{eig}


\begin{thebibliography}{51}


\ifx \showCODEN    \undefined \def \showCODEN     #1{\unskip}     \fi
\ifx \showDOI      \undefined \def \showDOI       #1{#1}\fi
\ifx \showISBNx    \undefined \def \showISBNx     #1{\unskip}     \fi
\ifx \showISBNxiii \undefined \def \showISBNxiii  #1{\unskip}     \fi
\ifx \showISSN     \undefined \def \showISSN      #1{\unskip}     \fi
\ifx \showLCCN     \undefined \def \showLCCN      #1{\unskip}     \fi
\ifx \shownote     \undefined \def \shownote      #1{#1}          \fi
\ifx \showarticletitle \undefined \def \showarticletitle #1{#1}   \fi
\ifx \showURL      \undefined \def \showURL       {\relax}        \fi
\providecommand\bibfield[2]{#2}
\providecommand\bibinfo[2]{#2}
\providecommand\natexlab[1]{#1}
\providecommand\showeprint[2][]{arXiv:#2}

\bibitem[\protect\citeauthoryear{Achanta, Hemami, Estrada, and
  S{\"u}sstrunk}{Achanta et~al\mbox{.}}{2009}]%
        {achanta2009frequency}
\bibfield{author}{\bibinfo{person}{Radhakrishna Achanta},
  \bibinfo{person}{Sheila Hemami}, \bibinfo{person}{Francisco Estrada}, {and}
  \bibinfo{person}{Sabine S{\"u}sstrunk}.} \bibinfo{year}{2009}\natexlab{}.
\newblock \showarticletitle{Frequency-tuned salient region detection}. In
  \bibinfo{booktitle}{\emph{Proceedings of the IEEE Conference on Computer
  Vision and Pattern Recognition}}.
\newblock


\bibitem[\protect\citeauthoryear{Borji, Cheng, Jiang, and Li}{Borji
  et~al\mbox{.}}{2015}]%
        {borji2015salient}
\bibfield{author}{\bibinfo{person}{Ali Borji}, \bibinfo{person}{Ming-Ming
  Cheng}, \bibinfo{person}{Huaizu Jiang}, {and} \bibinfo{person}{Jia Li}.}
  \bibinfo{year}{2015}\natexlab{}.
\newblock \showarticletitle{Salient object detection: A benchmark}.
\newblock \bibinfo{journal}{\emph{IEEE Transactions on Image Processing}}
  \bibinfo{volume}{24}, \bibinfo{number}{12} (\bibinfo{year}{2015}),
  \bibinfo{pages}{5706--5722}.
\newblock


\bibitem[\protect\citeauthoryear{Bruno, Gugliuzza, Pirrone, and
  Ardizzone}{Bruno et~al\mbox{.}}{2020}]%
        {bruno2020multi}
\bibfield{author}{\bibinfo{person}{Alessandro Bruno},
  \bibinfo{person}{Francesco Gugliuzza}, \bibinfo{person}{Roberto Pirrone},
  {and} \bibinfo{person}{Edoardo Ardizzone}.} \bibinfo{year}{2020}\natexlab{}.
\newblock \showarticletitle{A Multi-Scale Colour and Keypoint Density-Based
  Approach for Visual Saliency Detection}.
\newblock \bibinfo{journal}{\emph{IEEE Access}}  \bibinfo{volume}{8}
  (\bibinfo{year}{2020}), \bibinfo{pages}{121330--121343}.
\newblock


\bibitem[\protect\citeauthoryear{Chen, Tan, Wang, and Hu}{Chen
  et~al\mbox{.}}{2018}]%
        {chen2018reverse}
\bibfield{author}{\bibinfo{person}{Shuhan Chen}, \bibinfo{person}{Xiuli Tan},
  \bibinfo{person}{Ben Wang}, {and} \bibinfo{person}{Xuelong Hu}.}
  \bibinfo{year}{2018}\natexlab{}.
\newblock \showarticletitle{Reverse attention for salient object detection}. In
  \bibinfo{booktitle}{\emph{European Conference on Computer Vision}}.
\newblock


\bibitem[\protect\citeauthoryear{Cheng, Mitra, Huang, and Hu}{Cheng
  et~al\mbox{.}}{2014a}]%
        {cheng2014salientshape}
\bibfield{author}{\bibinfo{person}{Ming-Ming Cheng}, \bibinfo{person}{Niloy~J
  Mitra}, \bibinfo{person}{Xiaolei Huang}, {and} \bibinfo{person}{Shi-Min Hu}.}
  \bibinfo{year}{2014}\natexlab{a}.
\newblock \showarticletitle{Salientshape: group saliency in image collections}.
\newblock \bibinfo{journal}{\emph{The Visual Computer}} \bibinfo{volume}{30},
  \bibinfo{number}{4} (\bibinfo{year}{2014}), \bibinfo{pages}{443--453}.
\newblock


\bibitem[\protect\citeauthoryear{Cheng, Mitra, Huang, Torr, and Hu}{Cheng
  et~al\mbox{.}}{2014b}]%
        {cheng2014global}
\bibfield{author}{\bibinfo{person}{Ming-Ming Cheng}, \bibinfo{person}{Niloy~J
  Mitra}, \bibinfo{person}{Xiaolei Huang}, \bibinfo{person}{Philip~HS Torr},
  {and} \bibinfo{person}{Shi-Min Hu}.} \bibinfo{year}{2014}\natexlab{b}.
\newblock \showarticletitle{Global contrast based salient region detection}.
\newblock \bibinfo{journal}{\emph{IEEE Transactions on Pattern Analysis and
  Machine Intelligence}} \bibinfo{volume}{37}, \bibinfo{number}{3}
  (\bibinfo{year}{2014}), \bibinfo{pages}{569--582}.
\newblock


\bibitem[\protect\citeauthoryear{Deng, Dong, Socher, Li, Li, and Fei-Fei}{Deng
  et~al\mbox{.}}{2009}]%
        {deng2009imagenet}
\bibfield{author}{\bibinfo{person}{Jia Deng}, \bibinfo{person}{Wei Dong},
  \bibinfo{person}{Richard Socher}, \bibinfo{person}{Li-Jia Li},
  \bibinfo{person}{Kai Li}, {and} \bibinfo{person}{Li Fei-Fei}.}
  \bibinfo{year}{2009}\natexlab{}.
\newblock \showarticletitle{Imagenet: A large-scale hierarchical image
  database}. In \bibinfo{booktitle}{\emph{Proceedings of the IEEE Conference on
  Computer Vision and Pattern Recognition}}.
\newblock


\bibitem[\protect\citeauthoryear{Everingham, Van~Gool, Williams, Winn, and
  Zisserman}{Everingham et~al\mbox{.}}{2010}]%
        {everingham2010pascal}
\bibfield{author}{\bibinfo{person}{Mark Everingham}, \bibinfo{person}{Luc
  Van~Gool}, \bibinfo{person}{Christopher~KI Williams}, \bibinfo{person}{John
  Winn}, {and} \bibinfo{person}{Andrew Zisserman}.}
  \bibinfo{year}{2010}\natexlab{}.
\newblock \showarticletitle{The pascal visual object classes (voc) challenge}.
\newblock \bibinfo{journal}{\emph{Springer International Journal of Computer
  Vision}} \bibinfo{volume}{88}, \bibinfo{number}{2} (\bibinfo{year}{2010}),
  \bibinfo{pages}{303--338}.
\newblock


\bibitem[\protect\citeauthoryear{Fan, Cheng, Liu, Li, and Borji}{Fan
  et~al\mbox{.}}{2017}]%
        {fan2017structure}
\bibfield{author}{\bibinfo{person}{Deng-Ping Fan}, \bibinfo{person}{Ming-Ming
  Cheng}, \bibinfo{person}{Yun Liu}, \bibinfo{person}{Tao Li}, {and}
  \bibinfo{person}{Ali Borji}.} \bibinfo{year}{2017}\natexlab{}.
\newblock \showarticletitle{Structure-measure: A new way to evaluate foreground
  maps}. In \bibinfo{booktitle}{\emph{Proceedings of the IEEE/CVF International
  Conference on Computer Vision}}.
\newblock


\bibitem[\protect\citeauthoryear{Fan, Gong, Cao, Ren, Cheng, and Borji}{Fan
  et~al\mbox{.}}{2018}]%
        {fan2018enhanced}
\bibfield{author}{\bibinfo{person}{Deng-Ping Fan}, \bibinfo{person}{Cheng
  Gong}, \bibinfo{person}{Yang Cao}, \bibinfo{person}{Bo Ren},
  \bibinfo{person}{Ming-Ming Cheng}, {and} \bibinfo{person}{Ali Borji}.}
  \bibinfo{year}{2018}\natexlab{}.
\newblock \showarticletitle{Enhanced-alignment measure for binary foreground
  map evaluation}. In \bibinfo{booktitle}{\emph{IJCAI}}.
\newblock


\bibitem[\protect\citeauthoryear{Felzenszwalb and Huttenlocher}{Felzenszwalb
  and Huttenlocher}{2004}]%
        {felzenszwalb2004efficient}
\bibfield{author}{\bibinfo{person}{Pedro~F Felzenszwalb} {and}
  \bibinfo{person}{Daniel~P Huttenlocher}.} \bibinfo{year}{2004}\natexlab{}.
\newblock \showarticletitle{Efficient graph-based image segmentation}.
\newblock \bibinfo{journal}{\emph{Springer INTERNATIONAL JOURNAL OF COMPUTER
  VISION}} \bibinfo{volume}{59}, \bibinfo{number}{2} (\bibinfo{year}{2004}),
  \bibinfo{pages}{167--181}.
\newblock


\bibitem[\protect\citeauthoryear{Feng, Lu, and Ding}{Feng
  et~al\mbox{.}}{2019}]%
        {feng2019attentive}
\bibfield{author}{\bibinfo{person}{Mengyang Feng}, \bibinfo{person}{Huchuan
  Lu}, {and} \bibinfo{person}{Errui Ding}.} \bibinfo{year}{2019}\natexlab{}.
\newblock \showarticletitle{Attentive feedback network for boundary-aware
  salient object detection}. In \bibinfo{booktitle}{\emph{Proceedings of the
  IEEE Conference on Computer Vision and Pattern Recognition}}.
\newblock


\bibitem[\protect\citeauthoryear{Guo, Ma, and Zhang}{Guo et~al\mbox{.}}{2008}]%
        {guo2008spatio}
\bibfield{author}{\bibinfo{person}{Chenlei Guo}, \bibinfo{person}{Qi Ma}, {and}
  \bibinfo{person}{Liming Zhang}.} \bibinfo{year}{2008}\natexlab{}.
\newblock \showarticletitle{Spatio-temporal saliency detection using phase
  spectrum of quaternion fourier transform}. In
  \bibinfo{booktitle}{\emph{Proceedings of the IEEE Conference on Computer
  Vision and Pattern Recognition}}.
\newblock


\bibitem[\protect\citeauthoryear{He, Zhang, Ren, and Sun}{He
  et~al\mbox{.}}{2016}]%
        {he2016deep}
\bibfield{author}{\bibinfo{person}{Kaiming He}, \bibinfo{person}{Xiangyu
  Zhang}, \bibinfo{person}{Shaoqing Ren}, {and} \bibinfo{person}{Jian Sun}.}
  \bibinfo{year}{2016}\natexlab{}.
\newblock \showarticletitle{Deep residual learning for image recognition}. In
  \bibinfo{booktitle}{\emph{Proceedings of the IEEE conference on computer
  vision and pattern recognition}}. \bibinfo{pages}{770--778}.
\newblock


\bibitem[\protect\citeauthoryear{Hou, Cheng, Hu, Borji, Tu, and Torr}{Hou
  et~al\mbox{.}}{2017}]%
        {hou2017deeply}
\bibfield{author}{\bibinfo{person}{Qibin Hou}, \bibinfo{person}{Ming-Ming
  Cheng}, \bibinfo{person}{Xiaowei Hu}, \bibinfo{person}{Ali Borji},
  \bibinfo{person}{Zhuowen Tu}, {and} \bibinfo{person}{Philip~HS Torr}.}
  \bibinfo{year}{2017}\natexlab{}.
\newblock \showarticletitle{Deeply supervised salient object detection with
  short connections}. In \bibinfo{booktitle}{\emph{Proceedings of the IEEE
  Conference on Computer Vision and Pattern Recognition}}.
\newblock


\bibitem[\protect\citeauthoryear{Hou and Zhang}{Hou and Zhang}{2007}]%
        {hou2007saliency}
\bibfield{author}{\bibinfo{person}{Xiaodi Hou} {and} \bibinfo{person}{Liqing
  Zhang}.} \bibinfo{year}{2007}\natexlab{}.
\newblock \showarticletitle{Saliency detection: A spectral residual approach}.
  In \bibinfo{booktitle}{\emph{Proceedings of the IEEE Conference on Computer
  Vision and Pattern Recognition}}.
\newblock


\bibitem[\protect\citeauthoryear{Li and Yu}{Li and Yu}{2016}]%
        {li2016visual}
\bibfield{author}{\bibinfo{person}{Guanbin Li} {and} \bibinfo{person}{Yizhou
  Yu}.} \bibinfo{year}{2016}\natexlab{}.
\newblock \showarticletitle{Visual saliency detection based on multiscale deep
  CNN features}.
\newblock \bibinfo{journal}{\emph{IEEE Transactions on Image Processing}}
  \bibinfo{volume}{25}, \bibinfo{number}{11} (\bibinfo{year}{2016}),
  \bibinfo{pages}{5012--5024}.
\newblock


\bibitem[\protect\citeauthoryear{Li, Hou, Koch, Rehg, and Yuille}{Li
  et~al\mbox{.}}{2014}]%
        {li2014secrets}
\bibfield{author}{\bibinfo{person}{Yin Li}, \bibinfo{person}{Xiaodi Hou},
  \bibinfo{person}{Christof Koch}, \bibinfo{person}{James~M Rehg}, {and}
  \bibinfo{person}{Alan~L Yuille}.} \bibinfo{year}{2014}\natexlab{}.
\newblock \showarticletitle{The secrets of salient object segmentation}. In
  \bibinfo{booktitle}{\emph{Proceedings of the IEEE Conference on Computer
  Vision and Pattern Recognition}}.
\newblock


\bibitem[\protect\citeauthoryear{Liu, Hou, Cheng, Feng, and Jiang}{Liu
  et~al\mbox{.}}{2019}]%
        {liu2019simple}
\bibfield{author}{\bibinfo{person}{Jiang-Jiang Liu}, \bibinfo{person}{Qibin
  Hou}, \bibinfo{person}{Ming-Ming Cheng}, \bibinfo{person}{Jiashi Feng}, {and}
  \bibinfo{person}{Jianmin Jiang}.} \bibinfo{year}{2019}\natexlab{}.
\newblock \showarticletitle{A simple pooling-based design for real-time salient
  object detection}. In \bibinfo{booktitle}{\emph{Proceedings of the IEEE
  Conference on Computer Vision and Pattern Recognition}}.
\newblock


\bibitem[\protect\citeauthoryear{Liu, Han, and Yang}{Liu et~al\mbox{.}}{2018}]%
        {liu2018picanet}
\bibfield{author}{\bibinfo{person}{Nian Liu}, \bibinfo{person}{Junwei Han},
  {and} \bibinfo{person}{Ming-Hsuan Yang}.} \bibinfo{year}{2018}\natexlab{}.
\newblock \showarticletitle{Picanet: Learning pixel-wise contextual attention
  for saliency detection}. In \bibinfo{booktitle}{\emph{Proceedings of the IEEE
  Conference on Computer Vision and Pattern Recognition}}.
\newblock


\bibitem[\protect\citeauthoryear{Long, Shelhamer, and Darrell}{Long
  et~al\mbox{.}}{2015}]%
        {long2015fully}
\bibfield{author}{\bibinfo{person}{Jonathan Long}, \bibinfo{person}{Evan
  Shelhamer}, {and} \bibinfo{person}{Trevor Darrell}.}
  \bibinfo{year}{2015}\natexlab{}.
\newblock \showarticletitle{Fully convolutional networks for semantic
  segmentation}. In \bibinfo{booktitle}{\emph{Proceedings of the IEEE
  Conference on Computer Vision and Pattern Recognition}}.
\newblock


\bibitem[\protect\citeauthoryear{Luo, Mishra, Achkar, Eichel, Li, and
  Jodoin}{Luo et~al\mbox{.}}{2017}]%
        {luo2017non}
\bibfield{author}{\bibinfo{person}{Zhiming Luo}, \bibinfo{person}{Akshaya
  Mishra}, \bibinfo{person}{Andrew Achkar}, \bibinfo{person}{Justin Eichel},
  \bibinfo{person}{Shaozi Li}, {and} \bibinfo{person}{Pierre-Marc Jodoin}.}
  \bibinfo{year}{2017}\natexlab{}.
\newblock \showarticletitle{Non-local deep features for salient object
  detection}. In \bibinfo{booktitle}{\emph{Proceedings of the IEEE Conference
  on Computer Vision and Pattern Recognition}}.
\newblock


\bibitem[\protect\citeauthoryear{Margolin, Zelnik-Manor, and Tal}{Margolin
  et~al\mbox{.}}{2014}]%
        {margolin2014evaluate}
\bibfield{author}{\bibinfo{person}{Ran Margolin}, \bibinfo{person}{Lihi
  Zelnik-Manor}, {and} \bibinfo{person}{Ayellet Tal}.}
  \bibinfo{year}{2014}\natexlab{}.
\newblock \showarticletitle{How to evaluate foreground maps?}. In
  \bibinfo{booktitle}{\emph{Proceedings of the IEEE Conference on Computer
  Vision and Pattern Recognition}}.
\newblock


\bibitem[\protect\citeauthoryear{M{\'a}ttyus, Luo, and Urtasun}{M{\'a}ttyus
  et~al\mbox{.}}{2017}]%
        {mattyus2017deeproadmapper}
\bibfield{author}{\bibinfo{person}{Gell{\'e}rt M{\'a}ttyus},
  \bibinfo{person}{Wenjie Luo}, {and} \bibinfo{person}{Raquel Urtasun}.}
  \bibinfo{year}{2017}\natexlab{}.
\newblock \showarticletitle{Deeproadmapper: Extracting road topology from
  aerial images}. In \bibinfo{booktitle}{\emph{Proceedings of the IEEE/CVF
  International Conference on Computer Vision}}.
\newblock


\bibitem[\protect\citeauthoryear{Pang, Zhao, Zhang, and Lu}{Pang
  et~al\mbox{.}}{2020}]%
        {pang2020multi}
\bibfield{author}{\bibinfo{person}{Youwei Pang}, \bibinfo{person}{Xiaoqi Zhao},
  \bibinfo{person}{Lihe Zhang}, {and} \bibinfo{person}{Huchuan Lu}.}
  \bibinfo{year}{2020}\natexlab{}.
\newblock \showarticletitle{Multi-Scale Interactive Network for Salient Object
  Detection}. In \bibinfo{booktitle}{\emph{Proceedings of the IEEE Conference
  on Computer Vision and Pattern Recognition}}.
\newblock


\bibitem[\protect\citeauthoryear{Perazzi, Kr{\"a}henb{\"u}hl, Pritch, and
  Hornung}{Perazzi et~al\mbox{.}}{2012}]%
        {perazzi2012saliency}
\bibfield{author}{\bibinfo{person}{Federico Perazzi}, \bibinfo{person}{Philipp
  Kr{\"a}henb{\"u}hl}, \bibinfo{person}{Yael Pritch}, {and}
  \bibinfo{person}{Alexander Hornung}.} \bibinfo{year}{2012}\natexlab{}.
\newblock \showarticletitle{Saliency filters: Contrast based filtering for
  salient region detection}. In \bibinfo{booktitle}{\emph{Proceedings of the
  IEEE Conference on Computer Vision and Pattern Recognition}}.
\newblock


\bibitem[\protect\citeauthoryear{Qin, Zhang, Huang, Gao, Dehghan, and
  Jagersand}{Qin et~al\mbox{.}}{2019}]%
        {qin2019basnet}
\bibfield{author}{\bibinfo{person}{Xuebin Qin}, \bibinfo{person}{Zichen Zhang},
  \bibinfo{person}{Chenyang Huang}, \bibinfo{person}{Chao Gao},
  \bibinfo{person}{Masood Dehghan}, {and} \bibinfo{person}{Martin Jagersand}.}
  \bibinfo{year}{2019}\natexlab{}.
\newblock \showarticletitle{Basnet: Boundary-aware salient object detection}.
  In \bibinfo{booktitle}{\emph{Proceedings of the IEEE Conference on Computer
  Vision and Pattern Recognition}}.
\newblock


\bibitem[\protect\citeauthoryear{Rahman and Wang}{Rahman and Wang}{2016}]%
        {rahman2016optimizing}
\bibfield{author}{\bibinfo{person}{Md~Atiqur Rahman} {and}
  \bibinfo{person}{Yang Wang}.} \bibinfo{year}{2016}\natexlab{}.
\newblock \showarticletitle{Optimizing intersection-over-union in deep neural
  networks for image segmentation}. In \bibinfo{booktitle}{\emph{International
  Symposium on Visual Computing}}.
\newblock


\bibitem[\protect\citeauthoryear{Su, Li, Zhang, Xia, and Tian}{Su
  et~al\mbox{.}}{2019}]%
        {su2019selectivity}
\bibfield{author}{\bibinfo{person}{Jinming Su}, \bibinfo{person}{Jia Li},
  \bibinfo{person}{Yu Zhang}, \bibinfo{person}{Changqun Xia}, {and}
  \bibinfo{person}{Yonghong Tian}.} \bibinfo{year}{2019}\natexlab{}.
\newblock \showarticletitle{Selectivity or invariance: Boundary-aware salient
  object detection}. In \bibinfo{booktitle}{\emph{Proceedings of the IEEE/CVF
  Conference on Computer Vision}}.
\newblock


\bibitem[\protect\citeauthoryear{Tong, Lu, Zhang, and Ruan}{Tong
  et~al\mbox{.}}{2015}]%
        {tong2015salient}
\bibfield{author}{\bibinfo{person}{Na Tong}, \bibinfo{person}{Huchuan Lu},
  \bibinfo{person}{Ying Zhang}, {and} \bibinfo{person}{Xiang Ruan}.}
  \bibinfo{year}{2015}\natexlab{}.
\newblock \showarticletitle{Salient object detection via global and local
  cues}.
\newblock \bibinfo{journal}{\emph{Pattern Recognition}} \bibinfo{volume}{48},
  \bibinfo{number}{10} (\bibinfo{year}{2015}), \bibinfo{pages}{3258--3267}.
\newblock


\bibitem[\protect\citeauthoryear{Treisman and Gelade}{Treisman and
  Gelade}{1980}]%
        {treisman1980feature}
\bibfield{author}{\bibinfo{person}{Anne~M Treisman} {and}
  \bibinfo{person}{Garry Gelade}.} \bibinfo{year}{1980}\natexlab{}.
\newblock \showarticletitle{A feature-integration theory of attention}.
\newblock \bibinfo{journal}{\emph{Cognitive Psychology}} \bibinfo{volume}{12},
  \bibinfo{number}{1} (\bibinfo{year}{1980}), \bibinfo{pages}{97--136}.
\newblock


\bibitem[\protect\citeauthoryear{Wang, Lu, Wang, Feng, Wang, Yin, and
  Ruan}{Wang et~al\mbox{.}}{2017}]%
        {wang2017learning}
\bibfield{author}{\bibinfo{person}{Lijun Wang}, \bibinfo{person}{Huchuan Lu},
  \bibinfo{person}{Yifan Wang}, \bibinfo{person}{Mengyang Feng},
  \bibinfo{person}{Dong Wang}, \bibinfo{person}{Baocai Yin}, {and}
  \bibinfo{person}{Xiang Ruan}.} \bibinfo{year}{2017}\natexlab{}.
\newblock \showarticletitle{Learning to detect salient objects with image-level
  supervision}. In \bibinfo{booktitle}{\emph{Proceedings of the IEEE Conference
  on Computer Vision and Pattern Recognition}}.
\newblock


\bibitem[\protect\citeauthoryear{Wang, Zhang, Wang, Lu, Yang, Ruan, and
  Borji}{Wang et~al\mbox{.}}{2018}]%
        {wang2018detect}
\bibfield{author}{\bibinfo{person}{Tiantian Wang}, \bibinfo{person}{Lihe
  Zhang}, \bibinfo{person}{Shuo Wang}, \bibinfo{person}{Huchuan Lu},
  \bibinfo{person}{Gang Yang}, \bibinfo{person}{Xiang Ruan}, {and}
  \bibinfo{person}{Ali Borji}.} \bibinfo{year}{2018}\natexlab{}.
\newblock \showarticletitle{Detect globally, refine locally: A novel approach
  to saliency detection}. In \bibinfo{booktitle}{\emph{Proceedings of the IEEE
  Conference on Computer Vision and Pattern Recognition}}.
\newblock


\bibitem[\protect\citeauthoryear{Wang, Lai, Fu, Shen, Ling, and Yang}{Wang
  et~al\mbox{.}}{2021}]%
        {wang2021salient}
\bibfield{author}{\bibinfo{person}{Wenguan Wang}, \bibinfo{person}{Qiuxia Lai},
  \bibinfo{person}{Huazhu Fu}, \bibinfo{person}{Jianbing Shen},
  \bibinfo{person}{Haibin Ling}, {and} \bibinfo{person}{Ruigang Yang}.}
  \bibinfo{year}{2021}\natexlab{}.
\newblock \showarticletitle{Salient object detection in the deep learning era:
  An in-depth survey}.
\newblock \bibinfo{journal}{\emph{IEEE Transactions on Pattern Analysis and
  Machine Intelligence}} (\bibinfo{year}{2021}).
\newblock


\bibitem[\protect\citeauthoryear{Wang, Simoncelli, and Bovik}{Wang
  et~al\mbox{.}}{2003}]%
        {wang2003multiscale}
\bibfield{author}{\bibinfo{person}{Zhou Wang}, \bibinfo{person}{Eero~P
  Simoncelli}, {and} \bibinfo{person}{Alan~C Bovik}.}
  \bibinfo{year}{2003}\natexlab{}.
\newblock \showarticletitle{Multiscale structural similarity for image quality
  assessment}. In \bibinfo{booktitle}{\emph{The Thrity-Seventh Asilomar
  Conference on Signals, Systems \& Computers}}.
\newblock


\bibitem[\protect\citeauthoryear{Wei, Wang, and Huang}{Wei
  et~al\mbox{.}}{2020a}]%
        {wei2020f3net}
\bibfield{author}{\bibinfo{person}{Jun Wei}, \bibinfo{person}{Shuhui Wang},
  {and} \bibinfo{person}{Qingming Huang}.} \bibinfo{year}{2020}\natexlab{a}.
\newblock \showarticletitle{F$^3$Net: Fusion, Feedback and Focus for Salient
  Object Detection}. In \bibinfo{booktitle}{\emph{Proceedings of the AAAI
  Conference on Artificial Intelligence}}.
\newblock


\bibitem[\protect\citeauthoryear{Wei, Wang, Wu, Su, Huang, and Tian}{Wei
  et~al\mbox{.}}{2020b}]%
        {wei2020label}
\bibfield{author}{\bibinfo{person}{Jun Wei}, \bibinfo{person}{Shuhui Wang},
  \bibinfo{person}{Zhe Wu}, \bibinfo{person}{Chi Su}, \bibinfo{person}{Qingming
  Huang}, {and} \bibinfo{person}{Qi Tian}.} \bibinfo{year}{2020}\natexlab{b}.
\newblock \showarticletitle{Label Decoupling Framework for Salient Object
  Detection}. In \bibinfo{booktitle}{\emph{Proceedings of the IEEE Conference
  on Computer Vision and Pattern Recognition}}.
\newblock


\bibitem[\protect\citeauthoryear{Wu, Feng, Guan, Wang, Lu, and Ding}{Wu
  et~al\mbox{.}}{2019a}]%
        {wu2019mutual}
\bibfield{author}{\bibinfo{person}{Runmin Wu}, \bibinfo{person}{Mengyang Feng},
  \bibinfo{person}{Wenlong Guan}, \bibinfo{person}{Dong Wang},
  \bibinfo{person}{Huchuan Lu}, {and} \bibinfo{person}{Errui Ding}.}
  \bibinfo{year}{2019}\natexlab{a}.
\newblock \showarticletitle{A mutual learning method for salient object
  detection with intertwined multi-supervision}. In
  \bibinfo{booktitle}{\emph{Proceedings of the IEEE Conference on Computer
  Vision and Pattern Recognition}}.
\newblock


\bibitem[\protect\citeauthoryear{Wu, Su, and Huang}{Wu et~al\mbox{.}}{2019b}]%
        {wu2019cascaded}
\bibfield{author}{\bibinfo{person}{Zhe Wu}, \bibinfo{person}{Li Su}, {and}
  \bibinfo{person}{Qingming Huang}.} \bibinfo{year}{2019}\natexlab{b}.
\newblock \showarticletitle{Cascaded partial decoder for fast and accurate
  salient object detection}. In \bibinfo{booktitle}{\emph{Proceedings of the
  IEEE Conference on Computer Vision and Pattern Recognition}}.
\newblock


\bibitem[\protect\citeauthoryear{Wu, Su, and Huang}{Wu et~al\mbox{.}}{2019c}]%
        {wu2019stacked}
\bibfield{author}{\bibinfo{person}{Zhe Wu}, \bibinfo{person}{Li Su}, {and}
  \bibinfo{person}{Qingming Huang}.} \bibinfo{year}{2019}\natexlab{c}.
\newblock \showarticletitle{Stacked cross refinement network for edge-aware
  salient object detection}. In \bibinfo{booktitle}{\emph{Proceedings of the
  IEEE/CVF International Conference on Computer Vision}}.
\newblock


\bibitem[\protect\citeauthoryear{Yan, Xu, Shi, and Jia}{Yan
  et~al\mbox{.}}{2013}]%
        {yan2013hierarchical}
\bibfield{author}{\bibinfo{person}{Qiong Yan}, \bibinfo{person}{Li Xu},
  \bibinfo{person}{Jianping Shi}, {and} \bibinfo{person}{Jiaya Jia}.}
  \bibinfo{year}{2013}\natexlab{}.
\newblock \showarticletitle{Hierarchical saliency detection}. In
  \bibinfo{booktitle}{\emph{Proceedings of the IEEE Conference on Computer
  Vision and Pattern Recognition}}.
\newblock


\bibitem[\protect\citeauthoryear{Yang, Zhang, Lu, Ruan, and Yang}{Yang
  et~al\mbox{.}}{2013}]%
        {yang2013saliency}
\bibfield{author}{\bibinfo{person}{Chuan Yang}, \bibinfo{person}{Lihe Zhang},
  \bibinfo{person}{Huchuan Lu}, \bibinfo{person}{Xiang Ruan}, {and}
  \bibinfo{person}{Ming-Hsuan Yang}.} \bibinfo{year}{2013}\natexlab{}.
\newblock \showarticletitle{Saliency detection via graph-based manifold
  ranking}. In \bibinfo{booktitle}{\emph{Proceedings of the IEEE Conference on
  Computer Vision and Pattern Recognition}}.
\newblock


\bibitem[\protect\citeauthoryear{Zhang, Dai, Lu, He, and Wang}{Zhang
  et~al\mbox{.}}{2018a}]%
        {zhang2018bi}
\bibfield{author}{\bibinfo{person}{Lu Zhang}, \bibinfo{person}{Ju Dai},
  \bibinfo{person}{Huchuan Lu}, \bibinfo{person}{You He}, {and}
  \bibinfo{person}{Gang Wang}.} \bibinfo{year}{2018}\natexlab{a}.
\newblock \showarticletitle{A bi-directional message passing model for salient
  object detection}. In \bibinfo{booktitle}{\emph{Proceedings of the IEEE
  Conference on Computer Vision and Pattern Recognition}}.
\newblock


\bibitem[\protect\citeauthoryear{Zhang, Wang, Lu, Wang, and Ruan}{Zhang
  et~al\mbox{.}}{2017b}]%
        {zhang2017amulet}
\bibfield{author}{\bibinfo{person}{Pingping Zhang}, \bibinfo{person}{Dong
  Wang}, \bibinfo{person}{Huchuan Lu}, \bibinfo{person}{Hongyu Wang}, {and}
  \bibinfo{person}{Xiang Ruan}.} \bibinfo{year}{2017}\natexlab{b}.
\newblock \showarticletitle{Amulet: Aggregating multi-level convolutional
  features for salient object detection}. In
  \bibinfo{booktitle}{\emph{Proceedings of the IEEE/CVF International
  Conference on Computer Vision}}.
\newblock


\bibitem[\protect\citeauthoryear{Zhang, Lin, Tao, Li, and Shi}{Zhang
  et~al\mbox{.}}{2017a}]%
        {zhang2017salient}
\bibfield{author}{\bibinfo{person}{Qing Zhang}, \bibinfo{person}{Jiajun Lin},
  \bibinfo{person}{Yanyun Tao}, \bibinfo{person}{Wenju Li}, {and}
  \bibinfo{person}{Yanjiao Shi}.} \bibinfo{year}{2017}\natexlab{a}.
\newblock \showarticletitle{Salient object detection via color and texture
  cues}.
\newblock \bibinfo{journal}{\emph{Neurocomputing}}  \bibinfo{volume}{243}
  (\bibinfo{year}{2017}), \bibinfo{pages}{35--48}.
\newblock


\bibitem[\protect\citeauthoryear{Zhang, Wang, Qi, Lu, and Wang}{Zhang
  et~al\mbox{.}}{2018b}]%
        {zhang2018progressive}
\bibfield{author}{\bibinfo{person}{Xiaoning Zhang}, \bibinfo{person}{Tiantian
  Wang}, \bibinfo{person}{Jinqing Qi}, \bibinfo{person}{Huchuan Lu}, {and}
  \bibinfo{person}{Gang Wang}.} \bibinfo{year}{2018}\natexlab{b}.
\newblock \showarticletitle{Progressive attention guided recurrent network for
  salient object detection}. In \bibinfo{booktitle}{\emph{Proceedings of the
  IEEE Conference on Computer Vision and Pattern Recognition}}.
\newblock


\bibitem[\protect\citeauthoryear{Zhao, Liu, Fan, Cao, Yang, and Cheng}{Zhao
  et~al\mbox{.}}{2019b}]%
        {zhao2019egnet}
\bibfield{author}{\bibinfo{person}{Jia-Xing Zhao}, \bibinfo{person}{Jiang-Jiang
  Liu}, \bibinfo{person}{Deng-Ping Fan}, \bibinfo{person}{Yang Cao},
  \bibinfo{person}{Jufeng Yang}, {and} \bibinfo{person}{Ming-Ming Cheng}.}
  \bibinfo{year}{2019}\natexlab{b}.
\newblock \showarticletitle{EGNet: Edge guidance network for salient object
  detection}. In \bibinfo{booktitle}{\emph{Proceedings of the IEEE/CVF
  Conference on Computer Vision}}.
\newblock


\bibitem[\protect\citeauthoryear{Zhao, Gao, Wang, and Cheng}{Zhao
  et~al\mbox{.}}{2019a}]%
        {zhao2019optimizing}
\bibfield{author}{\bibinfo{person}{Kai Zhao}, \bibinfo{person}{Shanghua Gao},
  \bibinfo{person}{Wenguan Wang}, {and} \bibinfo{person}{Ming-Ming Cheng}.}
  \bibinfo{year}{2019}\natexlab{a}.
\newblock \showarticletitle{Optimizing the f-measure for threshold-free salient
  object detection}. In \bibinfo{booktitle}{\emph{Proceedings of the IEEE/CVF
  International Conference on Computer Vision}}.
\newblock


\bibitem[\protect\citeauthoryear{Zhao and Wu}{Zhao and Wu}{2019}]%
        {zhao2019pyramid}
\bibfield{author}{\bibinfo{person}{Ting Zhao} {and} \bibinfo{person}{Xiangqian
  Wu}.} \bibinfo{year}{2019}\natexlab{}.
\newblock \showarticletitle{Pyramid feature attention network for saliency
  detection}. In \bibinfo{booktitle}{\emph{Proceedings of the IEEE Conference
  on Computer Vision and Pattern Recognition}}.
\newblock


\bibitem[\protect\citeauthoryear{Zhao, Pang, Zhang, Lu, and Zhang}{Zhao
  et~al\mbox{.}}{2020}]%
        {zhao2020suppress}
\bibfield{author}{\bibinfo{person}{Xiaoqi Zhao}, \bibinfo{person}{Youwei Pang},
  \bibinfo{person}{Lihe Zhang}, \bibinfo{person}{Huchuan Lu}, {and}
  \bibinfo{person}{Lei Zhang}.} \bibinfo{year}{2020}\natexlab{}.
\newblock \showarticletitle{Suppress and balance: A simple gated network for
  salient object detection}. In \bibinfo{booktitle}{\emph{European Conference
  on Computer Vision}}.
\newblock


\bibitem[\protect\citeauthoryear{Zhong, Li, Tang, Tang, and Ding}{Zhong
  et~al\mbox{.}}{2021}]%
        {zhong2021highly}
\bibfield{author}{\bibinfo{person}{Yijie Zhong}, \bibinfo{person}{Bo Li},
  \bibinfo{person}{Lv Tang}, \bibinfo{person}{Hao Tang}, {and}
  \bibinfo{person}{Shouhong Ding}.} \bibinfo{year}{2021}\natexlab{}.
\newblock \showarticletitle{Highly Efficient Natural Image Matting}. In
  \bibinfo{booktitle}{\emph{British Machine Vision Conference}}.
\newblock


\end{thebibliography}

\end{document}